%% file: main.tex
\documentclass[lettersize,journal]{IEEEtran}
\usepackage{amsmath,amsfonts}
\usepackage{algorithmic}
\usepackage{array}
\usepackage[caption=false,font=normalsize,labelfont=sf,textfont=sf]{subfig}
\usepackage{textcomp}
\usepackage{stfloats}
\usepackage{url}
\usepackage{verbatim}
\usepackage{graphicx}
\usepackage{array}
\usepackage{cuted}
\usepackage{caption}
\usepackage{multirow} 
\usepackage{makecell}
\usepackage{booktabs}
\usepackage{fontawesome}
\usepackage{pifont}
\usepackage{float}
\usepackage{color}
\newcommand{\cmark}{\ding{51}}%
\newcommand{\xmark}{\ding{55}}%

\usepackage{multirow}
\usepackage{makecell}
\usepackage{booktabs}
\usepackage{tabularx}
\usepackage{amssymb}
\usepackage{siunitx}
\usepackage{enumitem}
 
\def\BibTeX{{\rm B\kern-.05em{\sc i\kern-.025em b}\kern-.08em
    T\kern-.1667em\lower.7ex\hbox{E}\kern-.125emX}}
\usepackage{balance}
\begin{document}
\title{OmniVTA: Visuo-Tactile World Modeling for Contact-Rich \\ Robotic Manipulation}
\author{Anonymous Authors
}
\author{
Yuhang Zheng$^{1*}$,
Yujie Zang$^{1*}$,
Yupeng Zheng$^{2\dagger}$,
Weize Li$^{1}$,
Songen Gu$^{3}$,
Shuai Tian$^{2}$,
Xiang Li$^{4,5}$, \\
Ce Hao$^{6}$,
Chen Gao$^{1,7}$, 
Si Liu$^{7}$, 
Haoran Li$^{2}$,
Yilun Chen$^{4}$,
Shuicheng Yan$^{1\dagger}$,
Wenchao Ding$^{4\dagger}$ \\
\textsuperscript{1}National University of Singapore,
\textsuperscript{2}CASIA, 
\textsuperscript{3}Fudan University,
\textsuperscript{4}{TARS Robotics}, \\
\textsuperscript{5}Tsinghua University,
\textsuperscript{6}Zhongguancun Academy,
\textsuperscript{7}Beihang University \\
\textsuperscript{$\dagger$} Corresponding Author,
\textsuperscript{*} Equal Contribution \\
}
\maketitle
\begin{strip}
    \centering
    \includegraphics[width=\textwidth]{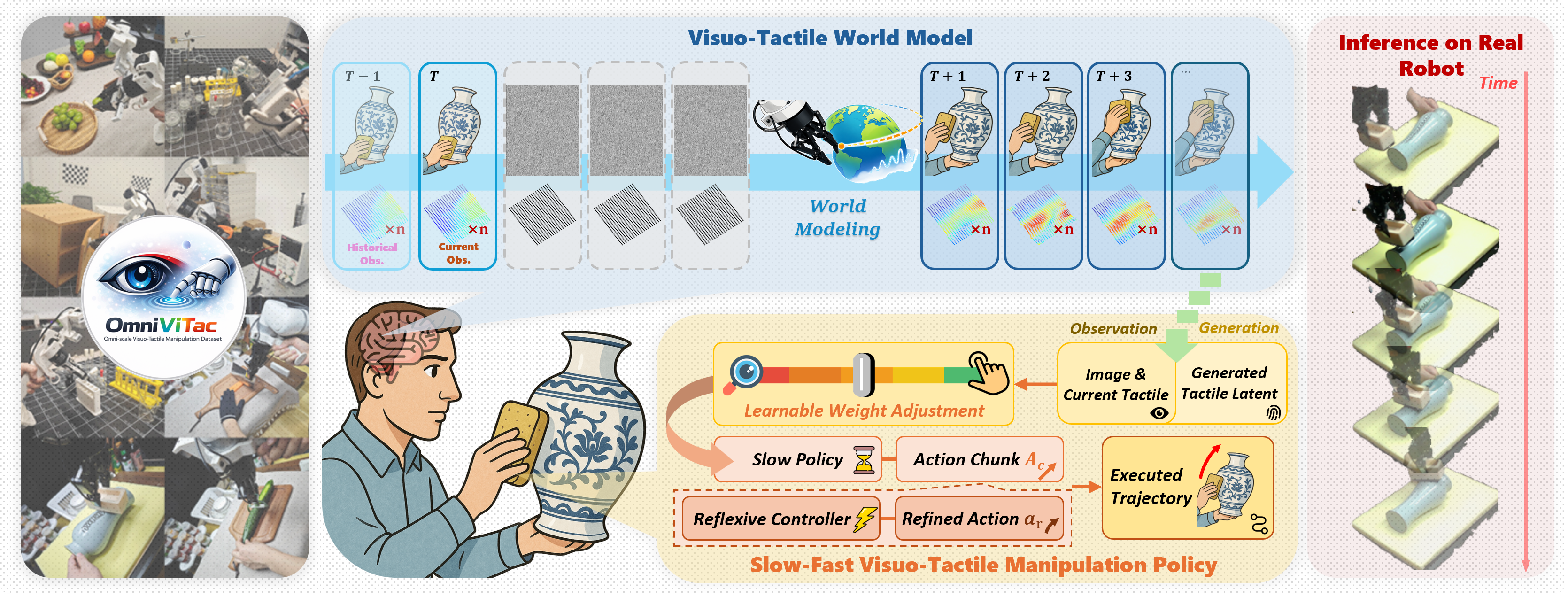}
    \captionof{figure}{\textcolor{black}{\textbf{Overview of the proposed visuo-tactile manipulation system.} (Left) We introduce OmniViTac, a large-scale visuo-tactile-action aligned dataset for contact-rich manipulation. (Center) We propose OmniVTA, a world model-based visuo-tactile-action framework that predicts future contact states. It seamlessly unifies tactile representation learning, predictive multimodal modeling, adaptive fusion, and reflexive tactile control. (Right) Extensive real-world experiments demonstrate that OmniVTA outperforms prior methods, exhibiting strong robustness and generalization.}}
    \label{fig:teaser}
\end{strip}

\begin{abstract}
Contact-rich manipulation tasks, such as wiping and assembly, require accurate perception of contact forces, friction changes, and state transitions that cannot be reliably inferred from vision alone. Despite growing interest in visuo-tactile manipulation, progress is constrained by two persistent limitations: existing datasets are small in scale and narrow in task coverage, and current methods treat tactile signals as passive observations rather than using them to model contact dynamics or enable closed-loop control explicitly.
In this paper, we present \textbf{OmniViTac}, a large-scale visuo-tactile-action dataset comprising $21{,}000+$ trajectories across $86$ tasks and $100+$ objects, organized into six physics-grounded interaction patterns. Building on this dataset, we propose \textbf{OmniVTA}, a world-model-based visuo-tactile manipulation framework that integrates four tightly coupled modules: a self-supervised tactile encoder, a two-stream visuo-tactile world model for predicting short-horizon contact evolution, a contact-aware fusion policy for action generation, and a 60Hz reflexive controller that corrects deviations between predicted and observed tactile signals in a closed loop.
Real-robot experiments across all six interaction categories show that OmniVTA outperforms existing methods and generalizes well to unseen objects and geometric configurations, confirming the value of combining predictive contact modeling with high-frequency tactile feedback for contact-rich manipulation. To advance research in visuo-tactile manipulation, all data, models, and code will be made publicly available at \url{https://mrsecant.github.io/OmniVTA}..
\end{abstract}


\input{sections/1_introduction}
\input{sections/2_related_work}
\input{sections/3_dataset}
\input{sections/4_method}
\input{sections/5_experiments}
\input{sections/6_conclusion}


\bibliographystyle{elsarticle-num} 
\bibliography{egbib}{}

\end{document}

%% file: sections/1_introduction.tex
\section{Introduction}
Contact-rich manipulation tasks, such as wiping and assembly, are ubiquitous in daily human activities. Their successful execution requires not only precise geometric alignment, but also accurate perception and fine-grained control of contact forces, frictional changes, and subtle transitions in contact state~\cite{zhao2024tac, cui2025vi, zhao2025tacman}. Since such information is difficult to infer reliably from vision alone, recent studies~\cite{bi2025vla, xue2025reactive, chen2025multi} have incorporated tactile sensing into imitation learning frameworks to improve visuomotor policies for contact-rich manipulation.

However, despite the growing adoption of tactile sensing in contact-rich manipulation, the broader visuo-tactile manipulation community still faces substantial limitations in both data and methodology. 
(1) On the data side, existing public visuo-tactile manipulation datasets~\cite{bi2025vla, yu2025demonstrating, bu2025agibot} remain limited in both the scale of fully aligned vision-tactile-action demonstrations and the diversity of contact-rich tasks. In particular, the amount of synchronized visual and high-frequency tactile data is still insufficient for learning generalizable visuo-tactile representations and modeling contact dynamics. In addition, the covered scenarios, tasks, and interaction patterns are narrow, making it difficult to capture the broad spectrum of contact mechanisms required in real-world manipulation.
(2) On the methodology side, existing visuo-tactile manipulation approaches remain limited in both representation and control. In most prior works~\cite{huang20243d, wu2025canonical}, tactile signals are incorporated merely as auxiliary observations to the policy network, primarily for contact-state recognition or compensating for visual occlusion, rather than being used to model the evolution of contact dynamics explicitly. At the control level, these methods typically rely on action chunking to generate short action sequences that are then executed in an open-loop manner. Such a design makes it difficult to react promptly to rapid contact changes, including slippage, misalignment, and external disturbances, which are pervasive in contact-rich manipulation. 

In contrast, humans perform such tasks with remarkable dexterity and robustness. Neuroscience studies~\cite{monzee2003effects, wolpert2001motor, kilteni2017sensorimotor, augurelle2003importance} suggest that this ability relies on tightly coupled multi-modal perception, predictive internal models, and rapid tactile feedback control. This combination of predictive modeling and reflexive correction is still largely missing from current visuo-tactile policies.

In this paper, we address these limitations from both the data and methodology perspectives. Concretely, we introduce OmniViTac, a large-scale visuo-tactile-action aligned dataset.
Building on this dataset, we further propose OmniVTA, a world-model-based visuo-tactile manipulation framework.

Specifically, OmniViTac is a large-scale visuo-tactile-action aligned dataset for contact-rich manipulation, comprising $21,879$ trajectories spanning $86$ tasks and $100+$ manipulated objects in diverse real-world scenarios. To capture the broad spectrum of contact-rich behaviors encountered in manipulation, we organize the dataset into six physics-grounded visuo-tactile interaction patterns: wiping, peeling, cutting, grasping, assembly, and in-hand adjustment.
To ensure data fidelity, OmniViTac is collected using a unified cross-embodiment platform and processed through a rigorous pipeline that preserves native sensor frequencies, performs precise temporal synchronization across vision, touch, and action streams, and incorporates automated trimming together with human-in-the-loop verification. Beyond scale and data quality, our analysis further shows that OmniViTac captures structured, pattern-specific contact dynamics and reveals two key properties of tactile signals in manipulation, namely spatial locality and contact-driven dynamics. 
These properties make OmniViTac a strong benchmark for learning visuo-tactile representations, predictive contact models, and robust manipulation policies.

Inspired by human sensorimotor control in contact-rich manipulation, an effective robotic system should possess two key capabilities~\cite{wolpert2001motor}: (1) forming feedforward anticipations of contact evolution through predictive modeling, and (2) exploiting tactile feedback for high-frequency closed-loop correction.

To this end, we propose OmniVTA, a world-model-based visuo-tactile manipulation framework composed of four tightly coupled modules. First, TactileVAE learns compact, low-dimensional tactile representations through spatio-temporal encoding and reconstructs continuous tactile deformation fields with an implicit neural decoder, providing structured tactile features for downstream prediction and policy learning. Second, the Visuo-Tactile World Model adopts a two-stream conditional generative architecture to model the temporal evolution of visual observations and tactile signals, thereby forming short-horizon feedforward predictions of contact dynamics. Third, the Adaptive Visuo-Tactile Fusion Policy leverages a Latent Tactile Differential (LTD) Encoder to capture the relationship between current tactile observations and predicted future tactile features, and adaptively fuses visual and tactile information through a contact-aware gating mechanism for action generation. Finally, the Reflexive Latent Tactile Controller uses predicted and observed tactile features to produce fine-grained corrective actions at 60,Hz, enabling rapid closed-loop correction during manipulation.

Extensive real-robot experiments across six tactile interaction pattern categories validate the effectiveness of OmniVTA from three perspectives. First, OmniVTA achieves state-of-the-art performance on diverse contact-rich manipulation tasks, with strong advantages in object diversity, geometric generalization, and disturbance robustness. The full framework further outperforms its open-loop variant, highlighting the importance of high-frequency tactile feedback for stable contact control. Second, mechanistic analyses show that the policy adaptively reweights vision and touch according to the predicted contact state, and that accurate tactile prediction is crucial for reliable action generation. Third, OmniVTA maintains stable performance under unseen geometric configurations and unseen tools, suggesting that it learns transferable contact-relevant structure and dynamics rather than memorizing task-specific trajectories. Together, these results demonstrate the benefit of predictive visuo-tactile world modeling combined with rapid tactile-feedback-based correction.

The main contributions of this paper are three-fold:
(1) We present OmniViTac, a large-scale visuo-tactile-action aligned dataset for contact-rich manipulation spanning six tactile interaction pattern categories, and reveal two key properties of tactile signals: spatial locality and contact-driven dynamics.
(2) We propose OmniVTA, a world-model-based visuo-tactile manipulation framework that integrates tactile representation learning, predictive visuo-tactile world modeling, contact-aware fusion, and high-frequency closed-loop tactile control.
(3) Extensive real-world experiments show that OmniVTA outperforms state-of-the-art baselines across diverse contact-rich manipulation tasks, with strong robustness to perturbations and effective generalization to novel configurations and objects.

%% file: sections/2_related_work.tex
\begin{figure*}[ht]
\centering
\includegraphics[width=0.95\linewidth]{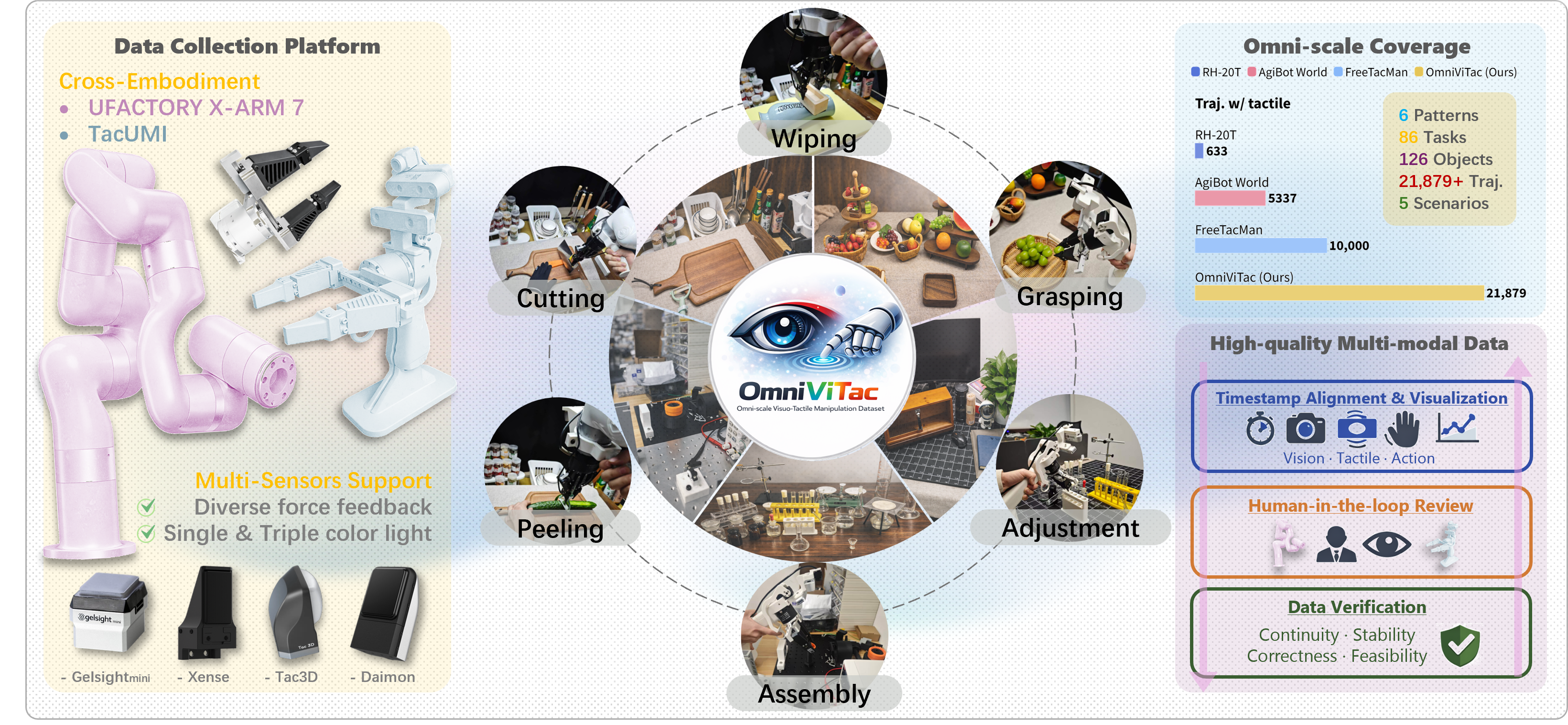}
\caption{{\textbf{{Overview of OmniViTac dataset.}} \textbf{Left:} The Cross-Embodiment Data Collection Platform features UNFactory 7-DoF xArm and the TacUMI manipulation interface, both supporting identical end-effectors and diverse tactile sensors (Xense, GelSight Mini, Tac3D, Daimon). \textbf{Middle:} The dataset covers 6 visuo-tactile manipulation patterns, instantiated across 5 semantic scenarios. \textbf{Top-Right:} A scale comparison demonstrates that OmniViTac (21,879 trajectories) significantly exceeds existing visuo-tactile manipulation datasets in tactile-rich data volume. \textbf{Bottom-Right:} The High-quality Data Pipeline ensures reliability through timestamp alignment, visualization, and human-in-the-loop verification.}}
\label{fig:dataset_teaser}
\end{figure*}

\section{Related Works}
\subsection{Tactile Sensing and Tactile Representation Learning}
Robotic fine-grained interaction relies on advances in both tactile hardware and representation learning. Existing tactile sensors span multiple physical principles, including capacitive arrays~\cite{xu2024cushsense,xu2016stretch}, piezoresistive sensors~\cite{stassi2014flexible, sundaram2019learning}, and magnetic sensors~\cite{yan2021soft}. Among them, visuo-tactile sensors such as GelSight~\cite{yuan10high} and DIGIT~\cite{lambeta2020digit} have attracted particular attention because they provide high-resolution and robust measurements. By tracking gel deformation with a camera, they reconstruct fine contact geometry and, with marker dots, estimate force and torque.

With such signals available, the focus has shifted from task-specific features to transferable tactile representations. A major line of work learns from paired vision-tactile data through self-supervision or contrastive learning. Sparsh~\cite{higuera2024sparsh} adopts a masked autoencoder, Anytouch~\cite{feng2025anytouch} uses contrastive learning, and UniT~\cite{xu2025unit} introduces a VQGAN-like architecture~\cite{esser2021taming} for compact latent modeling. More recently, tactile sensing has been integrated into MLLMs to align tactile, visual, and textual representations. Representative examples include Binding touch to everything~\cite{yang2024binding}, Octopi~\cite{yu2024octopi}, and VTV-LLM~\cite{xie2025universal}. Octopi-1.5~\cite{yu2025demonstrating} further adds retrieval augmentation to improve predictions for unseen object-tactile pairs. Our study builds on this trend by examining unified visuo-tactile representation learning across four visuo-tactile sensors: GelSight Mini~\cite{yuan10high}, Tac3D~\cite{zhang2022tac3d}, Xense~\cite{xense}, and Daimon~\cite{daimon}.

\subsection{Visuo-Tactile Robotic Manipulation Policies}
Findings from human motor control highlight the importance of tactile feedback for stable manipulation. Prior studies show that cutaneous feedback is essential for maintaining grip stability~\cite{augurelle2003importance} and that the brain uses a unified internal model to predict tactile feedback for both limbs and tools~\cite{kilteni2017sensorimotor, wolpert2001motor}. These results suggest that tactile sensing is fundamental to contact-rich manipulation, where it complements vision under occlusion and provides direct force-related feedback.

Inspired by these insights, robotic manipulation policies increasingly incorporate non-visual inputs. Early visuo-tactile systems such as See-to-Touch~\cite{guzey2024see} and RoboPack~\cite{ai2024robopack} showed that local tactile observations can resolve ambiguities and support fine-grained control when vision is unreliable. Recent work extends this idea to end-to-end manipulation models, including vision-action models such as 3D-ViTac~\cite{huang20243d} and RDP~\cite{xue2025reactive}, as well as vision-language-action models such as VLA-Touch~\cite{bi2025vla}, Tactile-VLA~\cite{huang2025tactile}, and TA-VLA~\cite{zhang2025ta}. However, effectively fusing tactile and visual signals for high-frequency closed-loop control remains difficult. Existing methods often suffer from weak tactile integration or temporal mismatch. T-Rex~\cite{niu2026trex} uses temporal tactile encoding and a variable-rate Mixture-of-Transformers for high-frequency feedback. Dream-Tac~\cite{lou2026dreamtac} jointly models actions and future visuo-tactile observations through contact-aware fusion. VTAM~\cite{yuan2026vtam} couples a pretrained video backbone with an action expert using a shared visual--tactile VAE.
Beyond direct policy learning, recent generative visuo-tactile modeling approaches predict interaction outcomes: VT-WM~\cite{Higuera2026VisuoTactileWM} learns action-conditioned visuo-tactile latent dynamics for model-based planning, whereas FlowTouch~\cite{Bien2026FlowTouchVV} generates
view-invariant static tactile observations from local 3D geometry. Different from these standalone generative or predictive models, our framework adopts a predictive--reactive slow--fast design, in which predicted future tactile latents support adaptive multimodal fusion and slow action generation, while real-time tactile feedback drives high-frequency residual correction for robust contact-rich manipulation.

\subsection{Visuo-tactile Manipulation Datasets and Systems}
Early visuo-tactile datasets~\cite{yang2024binding,fu2024touch,yu2024octopi,feng2025anytouch,xie2025universal,yu2025demonstrating} mainly collected static paired observations by pressing handheld tactile sensors against objects, surfaces, or textures. Although useful for multimodal representation learning, they do not provide action trajectories for manipulation. More recent datasets~\cite{li2022see,fang2023rh20t,wu2025freetacman,bi2025vla,cheng2025omnivtla,xu2025exumi,liu2025mla} mount tactile sensors on robotic grippers and record manipulation through UMI~\cite{chi2024universal}, teleoperation, or kinesthetic teaching. These systems capture synchronized vision, touch, and proprioception, enabling visuo-tactile policy learning for contact-rich tasks. Nevertheless, current datasets remain limited in scale, sensor and platform diversity, and temporal alignment between high-frequency tactile streams and visual observations.

Our work addresses these limitations with a multi-platform visuo-tactile manipulation dataset that offers diverse tasks and fully aligned vision, touch, and action sequences, supporting large-scale world-model training for unified visuo-tactile policy learning.

%% file: sections/3_dataset.tex




\section{The OmniViTac Dataset}


To support data-driven visuo-tactile learning for contact-rich manipulation, we introduce {OmniViTac}, a large-scale visuo-tactile-action dataset that addresses two key limitations of existing datasets: insufficient scale of fully aligned demonstrations and limited coverage of contact-rich tasks.

OmniViTac contains $21,879$ synchronized trajectories spanning $86$ tasks and $100+$ objects in diverse real-world environments, with RGB-D observations, high-frequency tactile sensing, and action streams recorded in a temporally aligned manner. 
Compared with prior visuo-tactile-action datasets in Tab.~\ref{tab:dataset_comparison}, OmniViTac substantially expands both task coverage and data scale, while also increasing sensor and collection diversity. Beyond scale alone, OmniViTac is organized around six physics-grounded visuo-tactile interaction patterns, enabling the dataset to capture a broad spectrum of contact dynamics rather than a narrow set of task-specific behaviors. 
These properties make OmniViTac a comprehensive benchmark for learning generalizable visuo-tactile representations, predictive contact models, and manipulation policies.

In the following, we describe OmniViTac in terms of its system design, task organization, and dataset scale. 
Sec.~\ref{sec:hardware} presents the hardware setup and synchronized data collection pipeline, which enables temporally aligned acquisition of RGB-D observations, tactile signals, and actions under diverse embodiments and sensing configurations. 
Sec.~\ref{sec:task} introduces the task taxonomy and representative visuo-tactile interaction patterns, revealing the contact-rich structure underlying the dataset. 
Sec.~\ref{sec:statistics} summarizes the dataset statistics and comparisons with prior work, highlighting the scale, diversity, and breadth of OmniViTac.

\begin{figure*}[ht]
\centering
\includegraphics[width=0.95\linewidth]{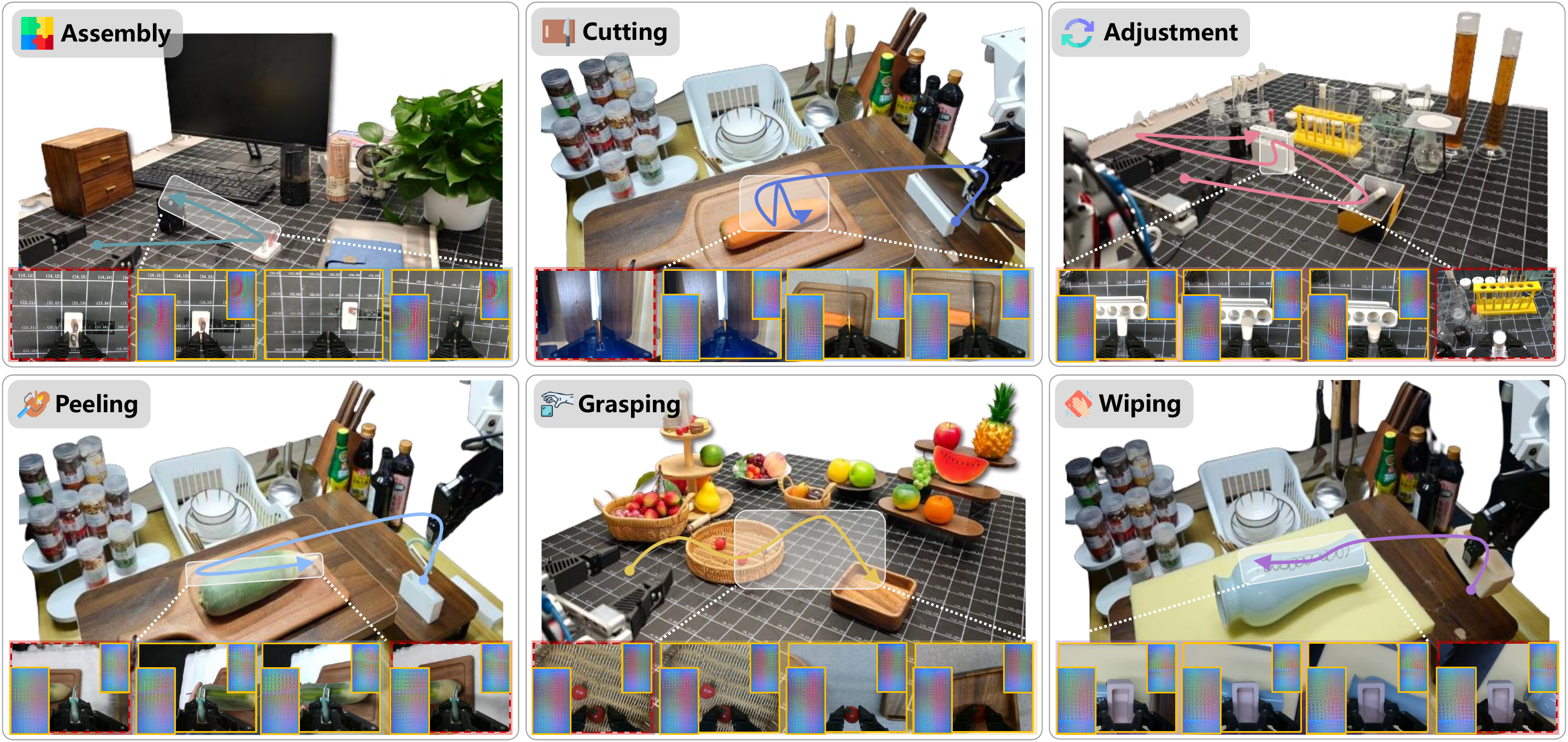}
\caption{\textbf{Example visualization of the $6$ visuo-tactile manipulation patterns in OmniViTac.} The dataset captures diverse, contact-rich behaviors across various scenarios, including Assembly, Cutting, Adjustment, Peeling, Grasping, and Wiping. Each panel illustrates the global third-person workspace view overlaid with the end-effector's trajectory, alongside synchronized, high-frequency tactile maps (bottom) that continuously record the complex tool-object contact dynamics during task execution.}
\label{fig:show_case}
\end{figure*}

\subsection{Hardware Setup and Data Collection System}
\label{sec:hardware}

\subsubsection{System Overview}

To efficiently collect large-scale visuo-tactile manipulation data, we develop a unified multimodal data acquisition system that can be operated by a single human operator. The system records synchronized multimodal observations during manipulation, including RGB-D visual observations, tactile signals from fingertip sensors, manipulation trajectories, and continuous gripper aperture states.

All sensory streams are recorded asynchronously at their native frequencies. A post-processing synchronization pipeline then aligns different modalities using timestamps to ensure consistent temporal correspondence across all observations.

\subsubsection{Dual-Embodiment Data Collection}

To balance \emph{robot-aligned demonstrations} and \emph{data collection scalability}, our dataset is collected using two complementary hardware embodiments: a 7-DoF robotic manipulator (xArm-7) and a TacUMI handheld data collection device. The robotic platform captures demonstrations that closely match the kinematic and dynamic distributions of robot execution, while TacUMI enables flexible and efficient human manipulation for large-scale data collection.

\noindent \textbf{Robot Demonstrations (7DoF Robot Arm).}
We employ a {UFACTORY xArm-7} manipulator to collect robot-aligned demonstrations. Two control modes are used:

\begin{itemize}
\item \textbf{Kinesthetic Teaching.} The operator directly guides the robot under gravity compensation mode, allowing precise control of contact interactions. This mode is particularly suitable for tasks requiring fine force perception, such as wiping surfaces or inserting connectors.

\item \textbf{GELLO Teleoperation.} For tasks involving larger spatial motions and less reliance on direct force perception, we adopt the {GELLO teleoperation system}~\cite{wu2024gello}. This master–follower interface enables efficient collection of pick-and-place style demonstrations while preserving realistic robot motion trajectories.
\end{itemize}

\begin{table*}[thb!]
    \centering
    \captionsetup{width=\textwidth}
    \caption{\textbf{Comparison of OmniViTac with existing vision-tactile-action datasets.} \textit{Note:} * indicates that the statistics count only the visuo-tactile trajectories in the corresponding dataset. ``--" denotes not applicable.
    } 
    \label{tab:dataset_comparison} 
    \vspace{-2mm}
    \addtolength{\tabcolsep}{-2pt}
    \resizebox{\textwidth}{!}{
        \begin{tabular}{@{}lccccccccc@{}}
            \toprule
            \multirow{2}{*}{\textbf{Dataset}} & \multirow{2}{*}{\textbf{\#Task}} & \multirow{2}{*}{\textbf{\#Traj.}} & \multirow{2}{*}{\textbf{\#Object}} & \multirow{2}{*}{\textbf{Vision}} & \multicolumn{3}{c}{\textbf{Tactile}} & \multirow{2}{*}{\textbf{Action}} & \multirow{2}{*}{\textbf{Collection}} \\ \cmidrule(lr){6-8}
             &  &  &  &  & \textbf{Sensors} & \textbf{Force Dim.} & \textbf{Frequency} &  &  \\ \midrule
            ObjectFolder2.0~\cite{gao2022objectfolder} & 3 & -- & 1,000 & Rendered & GelSight & 3 & -- & \xmark & Simulated \\
            AnyTouch~\cite{feng2025anytouch} & 1 & -- & 124 & -- & GelSight Mini/DIGIT/DuraGel/Tac3D & 3 & 30 & \xmark & Handheld \\
            VTV150k~\cite{xie2025universal} & -- & -- & 100 & -- & GelSight Mini/DIGIT/Tac3D & -- & -- & \xmark & Handheld \\
            Octopi-1.5~\cite{yu2025demonstrating} & 3 & -- & -- & GoPro & GelSight Mini/TAC-02 & -- & 50$\sim$1000 & \cmark & TMI \\
            See, Hear, and Feel~\cite{li2022see} & 2 & 60 & 4 & RealSense & GelSight & -- & 30 & \cmark & Teleop \\
            RH20T*~\cite{fang2023rh20t} & 55 & 917 & -- & RealSense & uSkin & -- & 200 & \cmark & Teleop \\
            FreeTacMan~\cite{wu2025freetacman} & 50 & 10,000 & 50 & Fisheye & FreeTacMan & -- & 30 & \cmark & Fingertips \\
            VLA-Touch~\cite{bi2025vla} & 3 & -- & -- & RealSense & GelSight Mini & -- & 10 & \cmark & Kinesthetic Teaching \\
            OmniVTLA~\cite{cheng2025omnivtla} & 2 & 240 & 56 & RealSense & GelSight/Paxini Gen 2 & -- & 30 & \xmark & Handheld \\
            exUMI~\cite{xu2025exumi} & 9 & 1,665 & 8 & GoPro & 9DTact & 6 & -- & \cmark & UMI-style \\
            MLA~\cite{liu2025mla} & 6 & 1,200 & 14 & RealSense & Tashan TS-E-A & 6 & -- & \xmark & Gello-teleop \\
            Hoi!~\cite{engelbracht2025hoi} & 6 & $\sim$3,000 & 381 & Zed Mini/Aria & DIGIT & 6 & 30$\sim$60 & \cmark & Gello-teleop \\
            AgiBot World*~\cite{bu2025agibot} & 7 & 5,337 & -- & -- & Xense & 3 & -- & \cmark & Teleop \\ \midrule
            \textbf{OmniViTac (Ours)} & 86 & 21,879 & 126 & RealSense/Fisheye & Xense/Gelsight Mini/DM-Tac/Tac3D & 3 & 30$\sim$60 & \cmark & UMI-style/Kinesthetic Teaching \\ \bottomrule
        \end{tabular}}
    \vspace{-5mm}
    \addtolength{\tabcolsep}{2pt}
\end{table*}

\noindent \textbf{Human Demonstrations (TacUMI).} 
To improve data collection efficiency and flexibility, we develop {TacUMI}, a handheld visuo-tactile data collection device inspired by FastUMI~\cite{zhaxizhuoma2025fastumi}. TacUMI enables natural manipulation motions while recording multimodal sensory observations that are consistent with the robotic platform. Following FastUMI, we use a {RealSense T265} tracking camera, which outputs 6-DoF poses at 200~Hz to estimate trajectories. After each trajectory is collected, an automatic verification step checks for tracking drift, and trajectories with position errors larger than 8~mm are discarded.

\noindent \textbf{Isomorphic End-Effector.}
To minimize embodiment-induced domain gaps, both the robotic arm and TacUMI employ identical parallel-jaw grippers. This shared end-effector design ensures consistent grasp geometry and contact mechanics across both data collection embodiments. Each gripper records \emph{continuous motor states}, which are normalized to a $[0,1]$ range to represent the gripper width. The fingertip structure is modular and supports interchangeable tactile sensor modules.

\subsubsection{Sensor Suite}
To capture the diverse perceptual cues required in contact-rich manipulation, we combine RGB-D sensing with multiple tactile sensors that provide complementary observations of object geometry, surface interaction, and contact dynamics.

\noindent \textbf{Visual Observation.}
We record both \emph{wrist-view} and \emph{third-person-view} RGB-D observations. Wrist-view images are captured using an Intel RealSense D435 camera, while third-person observations are captured using either a D435 or an L515 camera. All cameras record RGB-D streams at 30~Hz. The D435 provides images at $640\times480$ resolution, while the L515 provides $1280\times720$ resolution. TacUMI additionally integrates a wide-angle RGB camera to capture broader environmental context.

\noindent \textbf{Tactile Sensing.}
Tactile sensors are mounted on the gripper fingertips to capture detailed contact information during manipulation. Our platform supports four types of tactile sensors with different sensing principles and resolutions:

\begin{itemize}

\item \textbf{{Xense (Photon)}:}
records RGB tactile images with a resolution of $700\times400$ at 30~Hz and 3D displacement fields with a spatial resolution of $35\times20$ at 60~Hz.

\item \textbf{Daimon (Tac-WL):}
records grayscale tactile images at $640\times480$ resolution at 30~Hz and 3D displacement measurements with resolution $320\times240$ at 60~Hz.

\item \textbf{Tac3D-A1:}
provides dense 3D displacement sensing with spatial resolution $20\times20$ at 30~Hz.

\item \textbf{GelSight Mini:}
records RGB tactile images with resolution $320\times240$ and 3D displacement fields with spatial resolution $9\times7$, both captured at 25~Hz.

\end{itemize}

These sensors provide multiple tactile signal modalities for the same manipulation tasks. The Xense sensor is used for the majority of data collection due to its robustness in large-scale manipulation datasets. 
Consequently, demonstrations collected with Xense are primarily used for analysis and real-robot evaluation, while the additional tactile sensors are included to support research on tactile representation learning and cross-sensor generalization.

\noindent \textbf{Proprioceptive Signals.}
The system also records proprioceptive states, including robot joint states (only for xArm demonstrations), end-effector poses, and continuous gripper aperture values. During data collection, the average tactile displacement is visualized in real time to provide operators with an intuitive estimate of contact forces.

\subsubsection{Data Collection and Processing Pipeline}
We describe how visuo-tactile-action trajectories are recorded and how the raw data are collected, validated, and processed.

\noindent \textbf{Data Acquisition.}
To enable single-operator data collection, we design a foot-pedal control interface for managing the entire recording process. The interface provides three functions: starting the system, recording trajectories, and stopping it. During recording, all sensors continuously publish timestamped observations, which are subscribed and stored by the host computer. To prevent sensor drift from extended operation, the system automatically restarts after every 25 trajectories.

\noindent \textbf{Data Validation.}
We perform both online and offline quality checks. During collection, every 50 trajectories the system randomly selects three trajectories and visualizes synchronized sensor streams, including camera observations, tactile signals, trajectory curves, and gripper width signals. After data collection, all trajectories can be inspected offline using visualization tools, and abnormal samples are removed.

\noindent \textbf{Data Processing.}
Before policy training, we preprocess all trajectories to improve data quality and temporal consistency. This pipeline removes redundant static frames at the beginning and end of each trajectory, aligns RGB-D, tactile, and action streams through timestamp synchronization (with temporal error below 10~ms), and segments trajectories into training-ready clips according to downstream policy learning requirements.

\begin{figure*}[ht]
\centering
\includegraphics[width=0.95\linewidth]{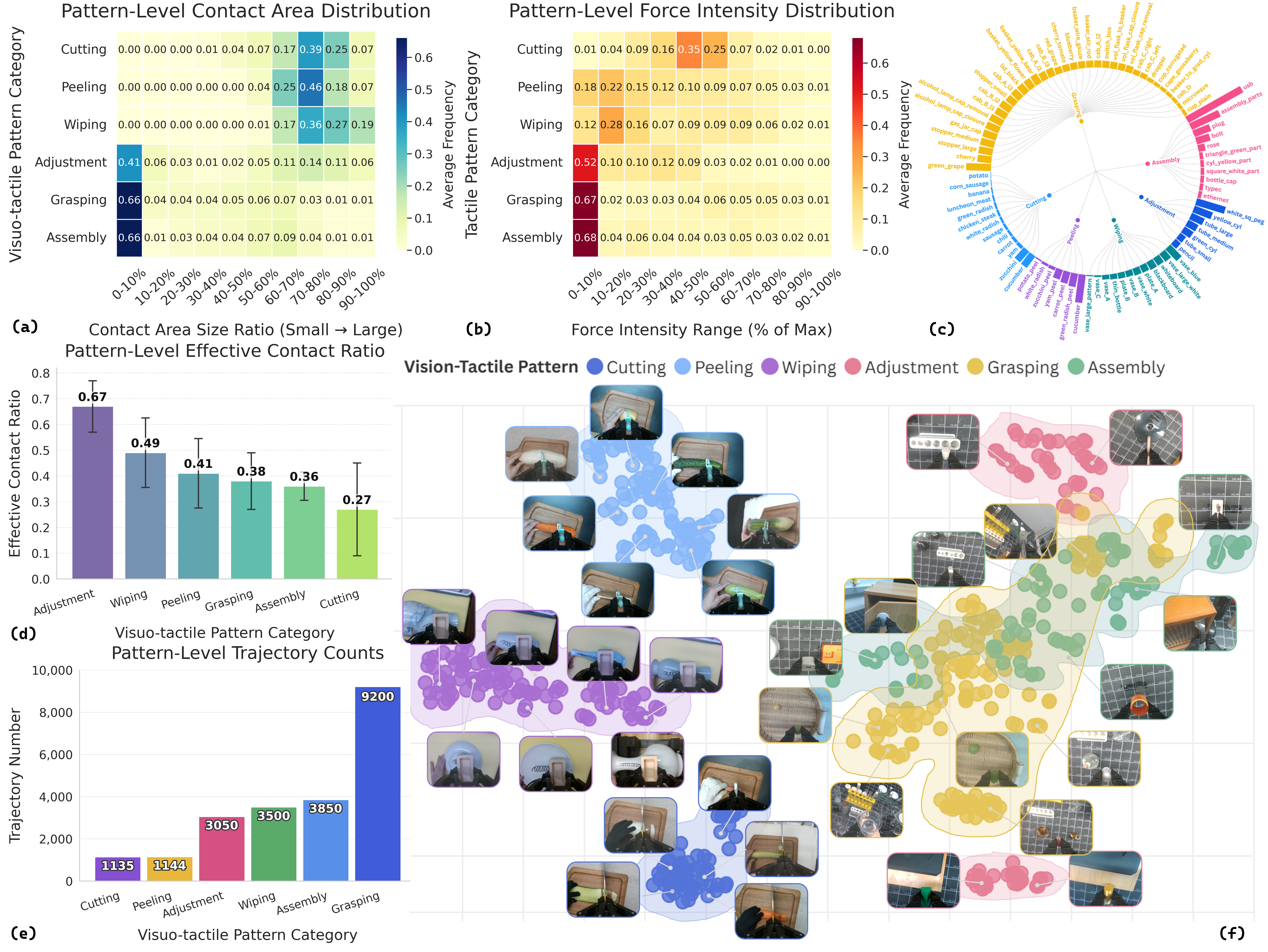}
\caption{\textbf{{Comprehensive statistical analysis of the OmniViTac.}} 
\textbf{(a)} Pattern-level contact area distribution, highlighting the distinct dichotomy between precision-dominant tasks (concentrated in the $0$-$10\%$ range) and surface-dominant tasks (peaking at $70$-$90\%$). 
\textbf{(b)} Force intensity distribution, showcasing the wide spectrum of force magnitudes required across different manipulation modes. 
\textbf{(c)} Hierarchical distribution illustrating the rich diversity of $86$ instantiated tasks. 
\textbf{(d)} Effective contact ratio with variance across patterns, demonstrating the temporal dependency on tactile feedback (e.g., \textit{Adjustment} exhibits the highest continuous engagement). 
\textbf{(e)} Total trajectory counts per category, demonstrating the massive scale and balanced composition of the benchmark. 
\textbf{(f)} t-SNE projection of the high-dimensional tactile signals, revealing physically intuitive and semantically separable latent clusters that strictly align with the underlying contact mechanics of each interaction pattern.}
\label{fig:statistic}
\end{figure*}

\subsection{Task Taxonomy and Tactile Pattern Analysis}
\label{sec:task}
\subsubsection{Environmental Diversity}

To facilitate robust generalization of visuo-tactile policies to real-world environments, OmniViTac is deliberately designed with high environmental diversity. As illustrated in the inner ring of Fig.~\ref{fig:dataset_teaser} and the examples in Fig.~\ref{fig:show_case}, the dataset encompasses over 70 distinct manipulation tasks, systematically distributed across five semantic scenarios: \textit{Kitchen, Fruit Shop, Industrial Workspace, Chemistry Laboratory,} and \textit{Office}. Diverse scene selection ensures broad coverage of visual distractors, varying illumination conditions, and a wide range of object properties, including rigid, deformable, transparent, and articulated types, thereby posing fundamental challenges to the robustness of policies.

\subsubsection{Task Taxonomy}
To move beyond task classification based purely on visual kinematics, we introduce a physics-grounded taxonomy that categorizes the collected behaviors into six representative visuo-tactile patterns. This taxonomy is defined by the dominant tactile features and contact mechanics underlying successful execution:
\begin{itemize}
    \item \textbf{Assembly:} Involves precise coordination of contact geometry and multi-directional forces to perceive tight tolerances and detect successful mating of intricate parts.
    \item \textbf{Cutting:} Predominantly relies on the progressive magnitude of normal forces to penetrate objects, using sudden force drops to judge the completion of the severing action.
    \item \textbf{Adjustment:} Characterized by torsional and shear forces, which are crucial for sensing slip and reorienting in-hand objects until the desired stable pose is achieved.
    \item \textbf{Peeling:} Involves a dynamic and continuous coupling of both shear and normal forces to maintain constant tool-surface contact while stripping object exteriors.
    \item \textbf{Wiping:} Another canonical contact-rich task demanding the simultaneous regulation of normal pressure (to maintain surface adherence) and planar shear force (to overcome friction during sweeping motions).
    \item \textbf{Grasping:} A highly versatile category encompassing a broad spectrum of force profiles. It utilizes precise normal force control to safely handle fragile items, leverages contact feedback to verify secure holds on visually challenging (e.g., transparent) objects, and exploits combined normal-shear dynamics to optimize the manipulation of complex articulated mechanisms.
\end{itemize}

\subsubsection{Tactile Pattern Analysis}

To quantitatively validate the proposed task taxonomy and analyze the underlying data distribution, we embed the high-dimensional tactile signals from all six pattern categories into a two-dimensional latent space using t-SNE, as shown in Fig.~\ref{fig:statistic}(f). The embedding reveals a highly structured and physically interpretable distribution. Notably, tasks dominated by continuous, high-frequency shear and dynamic normal forces, such as \textit{Wiping} and \textit{Peeling}, form overlapping or closely adjacent clusters, reflecting their shared underlying contact mechanics. In contrast, tasks governed primarily by static normal pressure and localized geometric contacts, such as \textit{Assembly}, yield distinct and well-separated manifolds. The \textit{Grasping} category occupies a broad and expansive region in the feature space, consistent with its inherent intra-class diversity spanning delicate handling of fragile objects to manipulation of complex articulated mechanisms. This clear semantic separability within the tactile latent space underscores that OmniViTac captures rich, pattern-specific contact dynamics rather than redundant or noisy signals.

\subsection{Dataset Statistics and Scale}
\label{sec:statistics}

\subsubsection{Overview Statistics}
To rigorously benchmark the scale, diversity, and fidelity of OmniViTac, we provide a comprehensive statistical analysis of all collected trajectories, with an emphasis on task distributions and their associated spatio-temporal tactile properties.
As illustrated in Fig.~\ref{fig:statistic}(e), the dataset captures a broad range of visuo-tactile patterns commonly encountered in daily manipulation tasks. The ``grasping" category constitutes the largest subset, comprising 9,200 trajectories. Other manipulation categories also account for a significant portion of the dataset, each comprising approximately 3,000 or 1,000 trajectories, with all categories remaining within the same order of magnitude. This balanced composition ensures sufficient data density for both general-purpose pre-training and task-specific fine-tuning.

\subsubsection{Tactile Active Ratio}

A key metric for evaluating contact-rich datasets is the Effective Contact Ratio, defined as the proportion of a trajectory during which the tactile sensors register meaningful tactile interaction. As shown in Fig.~\ref{fig:statistic}(d), \textit{Adjustment} exhibits the highest active ratio (0.67) with relatively low variance, primarily because in-hand reorientation demands continuous contact with the tool or object to monitor slip and state changes. Similarly, tasks such as \textit{Wiping} (0.49) and \textit{Peeling} (0.41) require prolonged frictional interaction. In contrast, \textit{Cutting} yields the lowest active ratio (0.27) and exhibits high variance, as these patterns typically involve an extended phase of visual alignment and pre-contact approach before the brief, high-force severing event.

\subsubsection{Pattern-level Contact Area and Force Intensity}

The heatmaps in Fig.~\ref{fig:statistic}(a) and (b) reveal a clear and physically intuitive distinction in how tactile sensors are spatially and dynamically engaged across different manipulation patterns.

\begin{itemize}
\item \textbf{Precision-oriented tasks} (e.g., \textit{Assembly, Grasping, Adjustment}): These manipulations are predominantly associated with the lowest deciles in both contact area (with $66\%$ of occurrences in the $0-10\%$ range) and force intensity. This distribution reflects the underlying nature of these tasks, which rely on delicate, localized fingertip contacts and sensitive low-magnitude force feedback to adjust object poses or prevent damage precisely.
\item \textbf{Power-oriented or surface-oriented tasks} (e.g., \textit{Cutting, Peeling, Wiping}): In contrast, these tasks exhibit extensive sensory activation. Their contact area distributions peak prominently in the $70\%-90\%$ range, indicating full-patch engagement of the sensor surface to maintain frictional contact. Moreover, \textit{Cutting} shows a marked shift toward higher force intensities (peaking in the $40\%-50\%$ range of sensor capacity), which is necessary for penetrating materials and overcoming physical resistance.
\end{itemize}

Together, these multi-dimensional statistics demonstrate that OmniViTac is not merely a collection of random contacts, but a highly structured and physically grounded benchmark, well-suited to support advanced research in visuo-tactile perception and contact-rich policy learning.

%% file: sections/4_method.tex
\begin{figure*}[ht]
\centering
\includegraphics[width=\linewidth]{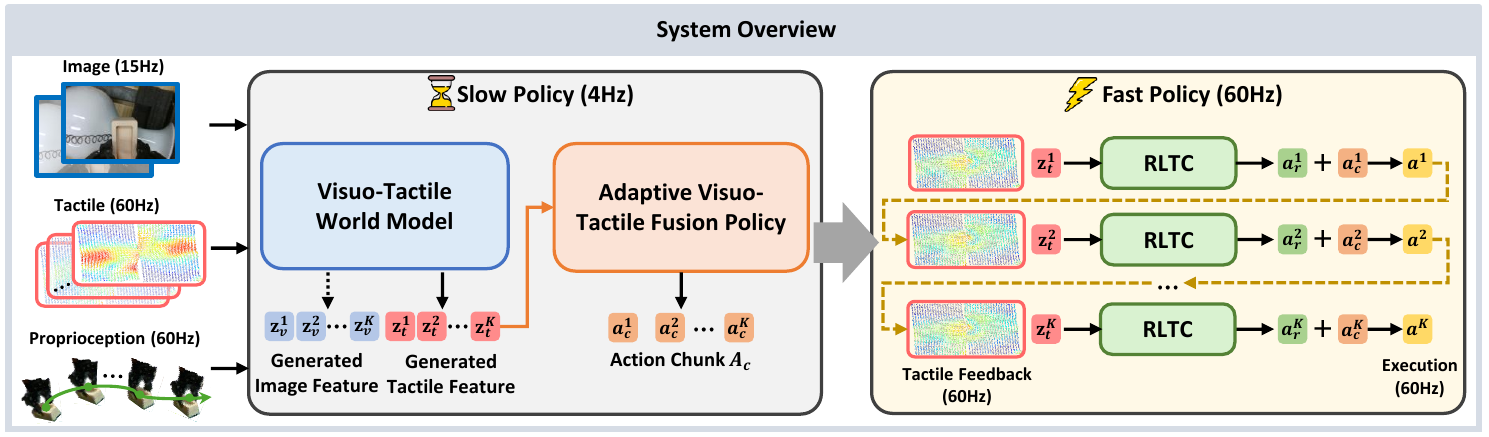}
\caption{\textbf{{System Overview}}. OmniVTA is a hierarchical slow–fast policy for contact-rich manipulation. The slow policy contains a visuo-tactile world model and an adaptive fusion policy to generate long-horizon action chunks from multi-modal inputs. 
The fast policy outputs high-frequency refinements at $60$~Hz using tactile feedback. 
Final actions are a weighted summation of slow-planned and fast-refined outputs, enabling both long-horizon planning and reactive control for robust manipulation.}
\label{fig:system}
\end{figure*}

\section{Methodology}

\subsection{System Overview}
To achieve stable and robust contact-rich manipulation, we introduce {OmniVTA}, a hierarchical slow-fast manipulation policy built upon a visuo-tactile world model, as illustrated in Fig.~\ref{fig:system}. OmniVTA explicitly decomposes manipulation into \emph{slow planning} and \emph{fast reflexive control}. 
The \emph{Slow Policy} integrates the Visuo-Tactile World Model (VTWM) and the Adaptive Visuo-Tactile Fusion Policy (AFP) to perform long-horizon action chunk planning from low-frequency visual observations, high-frequency tactile signals, and proprioceptive states. To improve efficiency, the world model predicts future tactile signals without generating visual observations during inference, enabling higher-frequency rollout while preserving contact-relevant dynamics.
The \emph{Fast Policy} is implemented as a Reflexive Latent Tactile Controller (RLTC), inspired by the reflexive control paradigm~\cite{xu2025aprebot, xu2025rebot}, which outputs fine-grained corrective actions at 60~Hz based on both predicted and observed tactile signals. This controller refines the action sequences generated by the slow policy by combining the planned actions with the controller outputs through a weighted summation, where the controller contribution is scaled by a predefined coefficient before execution on the robot.
This slow-fast decomposition enables both long-horizon planning and high-frequency reflexive control, which are critical for contact-rich manipulation.

In the following, we detail the components of OmniVTA. Sec.~\ref{sec:tacvae} introduces the TactileVAE for tactile feature extraction. Sec.~\ref{sec:vtwm} presents the Visuo-Tactile World Model, which captures contact dynamics via joint visual-tactile prediction. Sec.~\ref{sec:fusion_policy} describes the adaptive fusion policy with a gated mechanism~\cite{cao2023multi} for integrating visual and tactile features. Sec.~\ref{sec:controller} details the Reflexive Latent Tactile Controller.

\subsection{TactileVAE}
\label{sec:tacvae}
TactileVAE aims to rapidly extract low-dimensional, continuous, and task-agnostic tactile representations for downstream prediction and policy learning. For optical tactile sensors, instead of using high-resolution tactile images, we adopt \textit{3D marker displacement} as tactile input. This representation captures contact-induced surface deformation while maintaining significantly lower resolution, enabling efficient feature extraction and higher-frequency inference. The structure of the TactileVAE is shown in Fig.~\ref{fig:vae}, which contains a spatial-temporal encoder and an implicit deformation decoder.

\subsubsection{Encoder}
We construct a spatio-temporal encoder to extract compact tactile token representations. Since the markers on the tactile surface are uniformly distributed, a single-frame tactile observation can be represented as a tensor of size $H \times W \times 3$ where $H$ and $W$ denote the number of markers along the $y$- and $x$-axes, respectively, and the channel dimension corresponds to the marker displacement along the $x$, $y$, and $z$ directions.

As tactile sensing typically operates at a higher frequency than vision, we perform joint temporal and spatial compression to reduce the number of tactile tokens and align them with visual tokens in time. To this end, we adopt a \textit{3D convolution-based variational autoencoder (VAE)} and pretrain it on a large corpus of real tactile data to learn low-dimensional, task-agnostic latent features.

\begin{figure}[t]
\centering
\includegraphics[width=\linewidth]{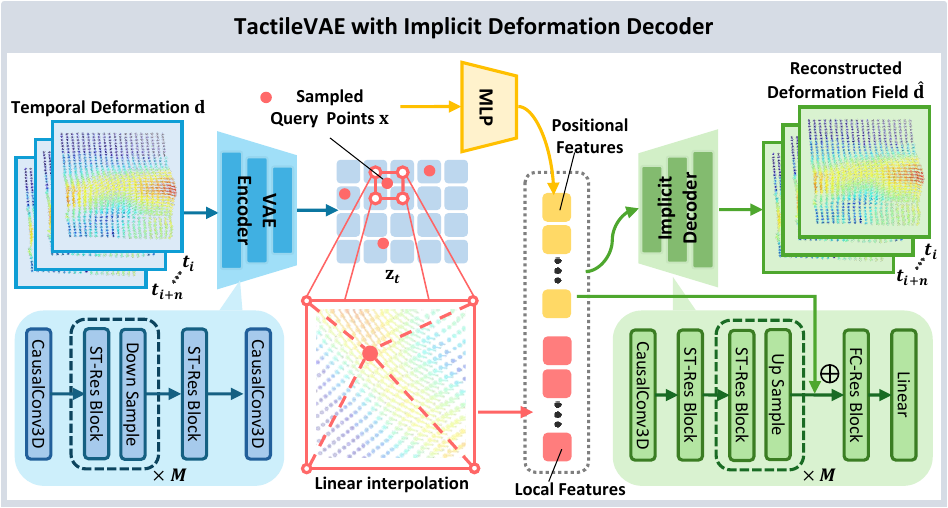}
\caption{\textbf{Overview of TactileVAE.} The model encodes 3D marker displacements into spatio-temporal features, and reconstructs the deformation via an implicit decoder that represents the deformation as a continuous field.}
\label{fig:vae}
\end{figure}

As illustrated in Fig.~\ref{fig:vae}, the encoder consists of a projection-in layer (causal 3D convolution), $M$ downsampling modules, and a projection-out layer (causal 3D convolution). The input is encoded into a latent feature map of size $\frac{H}{s} \times \frac{W}{s} \times C$, where $s = 2^M$ is the spatial downsampling factor and $C$ is the latent dimension. We adopt \textit{causal 3D convolutions} along the temporal axis so that the latent representation at time $t$ depends only on current and past observations, ensuring consistency between training and real-time deployment.

\subsubsection{Decoder}

Instead of reconstructing pixels as in conventional VAEs, we employ an \textit{implicit neural representation (INR)}~\cite{sitzmann2020implicit} decoder to model a continuous tactile deformation field conditioned on spatial coordinates. Implicit representations are well-suited for modeling continuous signals and have been widely used in computer vision tasks such as geometric reconstruction~\cite{zhong20233d, park2019deepsdf} and dense prediction~\cite{chen2023dpf, yu2026infinidepth}. Since marker displacement arises from the continuous deformation of the elastomer surface under contact, we model the deformation field as a continuous function:

\begin{equation}
    \mathbf{d}(\mathbf{x}) = \mathcal{D}_{\theta} \left( \gamma(\mathbf{x}), \Phi(\mathbf{z}_{t}, \mathbf{x}) \right),
\end{equation}
where $\mathbf{x} \in \mathbb{R}^2$ denotes the spatial coordinate, $\mathbf{z}_{t}$ is the latent feature map, $\gamma(\cdot)$ represents positional encoding, $\Phi(\mathbf{z}_{t},\mathbf{x})$ retrieves local features from $\mathbf{z}_{t}$ via spatial interpolation, and $\mathcal{D}_{\theta}$ is an MLP decoder that predicts the 3D deformation $\mathbf{d}(\mathbf{x}) \in \mathbb{R}^3$. 
During training, we uniformly sample query points in the latent feature map and extract their local features, which, together with the corresponding positional encodings, are fed into the decoder to predict the corresponding \textit{3D deformation vector}. This yields a continuous reconstruction of the tactile surface deformation field.

\subsubsection{Training Loss}
For supervision, the ground-truth deformation $\hat{\mathbf{d}}(\mathbf{x})$ at each query point $\mathbf{x}$ is obtained by interpolating the original 3D deformation. 
The TactileVAE is trained with a reconstruction loss and a KL regularization term~\cite{kingma2013auto}:

\begin{equation}
\mathcal{L}_{\text{TacVAE}} 
= \|\mathbf{d}(\mathbf{x}) - \hat{\mathbf{d}}(\mathbf{x})\|_2^2
+ \lambda_{\mathrm{KL}} \, \mathcal{L}_{\mathrm{KL}}.
\end{equation}

\begin{figure*}[ht]
\centering
\includegraphics[width=\linewidth]{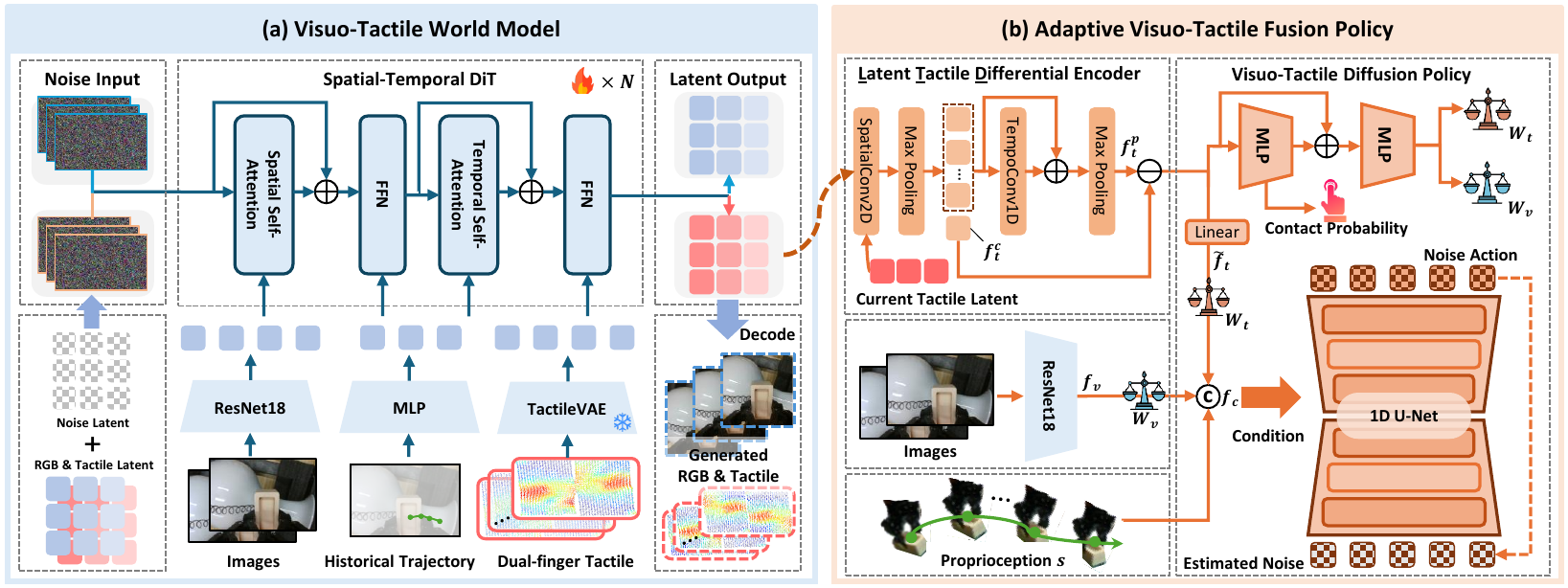}
\caption{\textbf{Overview of the Slow Policy.} \textbf{(a):} The proposed Visuo-Tactile World Model takes noisy tactile and visual latent sequences as input and jointly generates both modalities via a two-stream spatio-temporal diffusion transformer. \textbf{(b):} The Adaptive Visuo-Tactile Fusion Policy employs a Latent Tactile Differential Encoder to effectively encode the observed and predicted tactile and a gating mechanism to adaptively balance tactile and visual information for robust action planning.}
\label{fig:slow_policy}
\vspace{-5mm}
\end{figure*}

\vspace{-5mm}
\subsection{Visuo-Tactile World Model}
\label{sec:vtwm}
In this section, we present the Visuo-Tactile World Model, which adopts a two-stream conditional generative framework and explicitly models the temporal evolution of visual observations and tactile signals, thereby learning a dynamic prior for multimodal interactions. Overall, it consists of three key modules. The two-stream world model separately characterizes the dynamic generation process of the visual and tactile modalities while performing coordinated prediction under a shared condition. The multimodal observation conditioner jointly encodes vision, tactile, and actions, providing consistent conditioning for the generation process. The dynamic-aware weighted loss further emphasizes regions with high-frequency tactile variations and salient contact responses, improving the modeling of fine-grained tactile dynamics.

\subsubsection{Two-stream World Model}
To model the evolution of visual observations and tactile states during policy execution, we build a two-stream visuo-tactile world model, as shown in Fig.~\ref{fig:slow_policy} (a). Both branches employ a spatial-temporal diffusion transformer~\cite{peebles2023scalable}, which stacks transformer blocks along temporal and spatial axes. Specifically, the model takes the past $c$ frames as conditioning and iteratively denoises to generate multiple future frames, yielding a probabilistic model of future dynamics. The diffusion objective is defined as

\vspace{-2mm}
\begin{equation}
\mathcal{L}_{\text{diffusion}}=\mathbb{E}_{\mathbf{z}_o, \boldsymbol{\epsilon}, t}\left[\sum_{i=1}^K \left(1-m_i\right) \odot \left\|\epsilon_i-\boldsymbol{\epsilon}_\theta\left(\mathbf{z}_o, t\right)_i\right\|_2^2\right],
\end{equation}
where $\mathbf{z}_o = \left\{ \mathbf{z}_o^{1}, \dots, \mathbf{z}_o^{K} \right\}$ denotes a sequence of observation latents (including tactile latent $\mathbf{z}_t$ and visual latent $\mathbf{z}_v$), and $\boldsymbol{\epsilon}_\theta(\mathbf{z}_o, t)$ predicts the noise for the entire sequence. Each $\mathbf{z}_o^{i}$ corresponds to the encoded observation at timestep $i$. The mask $m$ provides temporal conditioning, encouraging the model to leverage historical observations for predicting future states.
At the modality encoding layer, the visual branch uses an SD-VAE~\cite{rombach2022high} to extract image latents, while the tactile branch employs the TactileVAE pretrained in the previous stage to compress tactile signals. After obtaining the latent representations of vision and touch, the two modalities are fed into their corresponding spatial-temporal diffusion models, enabling parallel modeling and joint generation. 

\subsubsection{Multi-modal Observation Conditioner}
To model joint visual-tactile dynamics effectively, we design a multi-modal conditional encoder for visual-tactile-action interactions. This module first performs feature extraction and temporal aggregation separately for visual observations, tactile embeddings, and {historical action sequences, where each action is represented as the 2D image-plane projection of the 3D end-effector position. Specifically, each historical 3D end-effector position is transformed from the robot base frame into the current camera frame and then projected onto the current image plane using the camera intrinsics. We observe that this representation improves robustness to variations in manipulation locations.} The modalities are then fused in a shared linear projection space to produce a fixed-dimensional conditioning vector. This vector is injected as a shared prior into both the tactile and visual world models, aligning the two branches and improving cross-modal consistency during dynamic prediction.

\subsubsection{Dynamic-aware Weighted Loss}
To supervise high-frequency dynamic changes in tactile signals more effectively, we introduce a Dynamic-Aware Weighted Loss. Specifically, the loss constructs a dynamic weight map based on local temporal differences of the tactile sequence to capture time-varying activity at different spatial locations:
    \begin{equation}
    w_{\text{dyn}^{i}} =
    \operatorname{resize}\left(
    \operatorname{clip}_{[0,1]}
    \left(
    \left\|X_{i+1}-X_i\right\|_2
    \right)
    \right),
    \end{equation}
where $X_k$ denotes the tactile frame at the $k$-th timestep, and $\operatorname{resize}(\cdot)$ resizes the weight map from the raw tactile resolution to the latent resolution. The dynamic-aware loss is defined as the weighted simple diffusion loss:
        \begin{equation}
        \mathcal{L}_{\text{dyn}} =
        \mathbb{E}_{\mathbf{z}_o,\boldsymbol{\epsilon},t}
        \left[
        \sum_{i=2}^{K}
        w_{\text{dyn}}^{i}\odot(1-m_i)
        \odot
        \left\|
        \epsilon_{i}-\boldsymbol{\epsilon}_\theta(\mathbf{z}_o,t)_{i}
        \right\|_2^2
        \right].
        \end{equation}
In addition, an amplitude weight map is constructed from the response magnitude of the tactile signal to represent the distribution of local contact intensity:
\begin{equation}
w_{\text{amp}^{i}} =
\operatorname{resize}\left(
\operatorname{clip}_{[0,1]}
\left(
\left\|X_i\right\|_2
\right)
\right).
\end{equation}
The amplitude loss is defined as
\begin{equation}
\mathcal{L}_{\text{amp}}=\mathbb{E}_{\boldsymbol{z_o}, \boldsymbol{\epsilon}, t}\left[\sum_{i=2}^K w_{\text{amp}}^i \odot \left(1-m_i\right) \odot \left\|\epsilon_i-\epsilon_\theta\left(\mathbf{z}_{o}^{i}, t\right)\right\|_2^2\right],
\end{equation}
 
The two weight maps are then aligned to the spatial resolution of the tactile latent representation and jointly used to modulate the reconstruction loss during training. This weighting strategy explicitly highlights regions with rapid dynamics and strong contact responses during optimization, thereby improving the modeling of high-frequency tactile patterns, local contact changes, and fine-grained temporal structure.
Finally, the overall objective is a weighted sum of the diffusion loss and the dynamic-aware weighted loss:
    \begin{equation}
    \mathcal{L}_{VTWM}
    =
    \mathcal{L}_{\text{diffusion}}
    +
    \lambda_1 \mathcal{L}_{\text{dyn}}
    +
    \lambda_2 \mathcal{L}_{\text{amp}},
    \end{equation}
where $\lambda_1$ and $\lambda_2$ are the loss weights for the dynamic and amplitude terms, respectively.

\subsection{Adaptive Visuo-Tactile Fusion Policy}
\label{sec:fusion_policy}
We propose the Adaptive Visuo-Tactile Fusion Policy, which adaptively balances visual and tactile information based on the predicted contact state for stable action planning. The policy comprises three components: a Latent Tactile Differential (LTD) Encoder for tactile representation, an Adaptive Visuo-Tactile Fusion module for modality weighting, and a Visuo-Tactile Diffusion Policy for action generation.

\subsubsection{Latent Tactile Differential (LTD) Encoder}
Tactile signals exhibit strong spatial locality and remain largely inactive before contact, becoming informative only when physical interaction occurs. Previous approaches typically concatenate historical tactile observations with visual features as policy inputs, which cannot explicitly capture potential contact dynamics and their consequences, limiting performance in contact-rich manipulation.

Inspired by the human feedforward prediction mechanism, we introduce a \textit{Latent Tactile Differential (LTD) Encoder} that captures potential contact states and interaction dynamics by modeling the difference between predicted future tactile features and the current tactile observation. Specifically, the current tactile feature map is spatially aggregated using 2D convolution and max pooling to obtain a global representation $f_t^c$. Predicted multi-frame tactile features are first aggregated spatially for each frame and then temporally using 1D convolution and max pooling to produce $f_t^p$. The final tactile representation is constructed through channel-wise concatenation of the current feature, the predicted feature, and their difference: 

\begin{equation}
f_t = \text{concat}(f_t^c, f_t^p, f_t^p - f_t^c),
\end{equation}
where $f_t^c$ is extracted from the tactile observation, while $f_t^p$ denotes the tactile latent predicted by the visuo-tactile world model. The differential term highlights discrepancies between predicted and current tactile signals, providing informative cues for potential contact events and interaction dynamics.

\subsubsection{Adaptive Visuo-Tactile Fusion}

To effectively integrate visual and tactile information, we design a gating-based adaptive fusion module that modulates modality contributions according to the interaction state.

Inspired by FoAR~\cite{he2025foar}, we first construct a contact probability prediction module to estimate the likelihood of physical contact occurring within a future time window. The predicted contact probability serves as a modulation signal for modality fusion. This module takes the tactile representation $f_t$ as input and predicts the contact probability using an MLP followed by a sigmoid activation. Contact labels are automatically generated by thresholding the tactile deformation magnitude: values exceeding a predefined threshold are labeled as contact, while others are treated as non-contact. The module is trained using BCE loss $\mathcal{L}_{bce}$.

Based on the predicted interaction state, we further construct a gating network consisting of two fully connected layers. The network takes the concatenation of the contact logit and tactile feature $f_t$ as input and outputs normalized modality weights $W_t$ and $W_v$, which modulate the tactile and visual features, respectively. $W_t$ and $W_v$ are per-channel weights normalized such that $W_t + W_v = 1$. Since the tactile representation already encodes future tactile dynamics through the world model, the gating network does not require visual inputs, reducing model complexity and improving computational efficiency. Before fusion, the tactile feature $f_t$ is projected to the same dimensionality as the visual feature using a linear layer, producing the projected tactile feature $\tilde{f}_t$. The fused visuo-tactile representation is therefore computed as

\begin{equation}
f_{vt} = \text{concat}\left(W_v \odot f_v,\; W_t \odot \tilde{f}_t \right),
\end{equation}
where $f_v$ denotes the visual feature extracted by a ResNet-18 encoder. Notably, we use only the current and historical visual observations without incorporating future visual prediction, since the current image already provides sufficient global context for action planning, while the predicted tactile features capture potential contact dynamics, making visual prediction unnecessary and reducing model complexity.

\subsubsection{Visuo-Tactile Diffusion Policy}
We model the visuo-tactile manipulation policy as a conditional denoising diffusion model~\cite{chi2025diffusion} conditioned on the fused visuo-tactile representation $f_{vt}$ and robot proprioception $s$. The policy generates a short-horizon action chunk by progressively denoising Gaussian noise. Specifically, the diffusion model predicts a sequence of coarse actions $A_c = (a_c^1, a_c^2, \dots, a_c^H)$, where $A_c$ denotes an action chunk consisting of $H$ coarse actions.

Starting from Gaussian noise $A_{c,T}$, the denoising network $\epsilon_\theta$ iteratively refines the action chunk through $T$ steps to obtain the clean action sequence $A_{c,0}$:

\begin{equation}
A_{c,t-1} = \alpha_t A_{c,t} - \gamma_k \epsilon_\theta(A_{c,t}, t, f_c) + \sigma_t \mathcal{N}(0, I),
\end{equation}

where $\mathcal{N}(0, I)$ denotes Gaussian noise and $\alpha_t$, $\gamma_t$, and $\sigma_t$ are scheduler-dependent coefficients. This reverse diffusion update resembles Stochastic Langevin Dynamics~\cite{welling2011bayesian}, where the noise predictor implicitly parameterizes the score function.
During training, the model is conditioned on $f_c = \text{concat}(f_{vt}, s)$ and optimized using the DDPM objective

\begin{equation}
\mathcal{L}_{act} =
\mathbb{E}_{t,\, A_{c,0}, \epsilon_t}
\left[
\left\|
\epsilon_t -
\epsilon_\theta
\left(
\bar{\alpha}_t A_{c,0} + \bar{\beta}_t \epsilon_t,\; t,\; f_c
\right)
\right\|_2^2
\right].
\end{equation}

To incorporate the conditional information, we adopt a CNN-based diffusion architecture where the conditioning feature $f_c$ is injected through FiLM modulation~\cite{perez2018film}.

\subsubsection{Training Loss}
The Adaptive Visuo-Tactile Fusion Policy is jointly optimized with an action loss $\mathcal{L}_{\text{act}}$ and a contact prediction loss $\mathcal{L}_{\text{bce}}$. The overall objective is defined as:

\begin{equation}
\mathcal{L}_{AFP}
=
\mathcal{L}_{\text{act}}
+
\lambda_{ct} \mathcal{L}_{\text{bce}}.
\end{equation}

\subsection{Reflexive Latent Tactile Controller}
\label{sec:controller}
Since the action chunk generated by the Visuo-Tactile Diffusion Policy is executed in an open-loop manner, we introduce a Reflexive Latent Tactile Controller to provide high-frequency tactile feedback and enable closed-loop control. 

\subsubsection{Model Design}
{The RLTC takes single-frame tactile feedback, predicted tactile features, and robot states as input and outputs a high-frequency refined action. Specifically, since the TactileVAE compresses tactile observations along the temporal dimension by a factor of $M=4$, the current single-frame tactile observation is temporally repeated four times to form a short sequence compatible with the encoder and is then encoded into a current tactile latent feature using the TactileVAE. Similarly, the tactile predictions generated by the world model have a lower temporal resolution of 15~Hz due to latent compression. We upsample the predicted tactile latent sequence using nearest-neighbor interpolation, such that each predicted latent is repeated and shared by four consecutive 60~Hz control steps, establishing an explicit 4-to-1 chunk-level temporal correspondence between the 15~Hz predicted tactile latents and the 60~Hz tactile feedback.}
The LTD Encoder is then applied to jointly encode the current tactile feature and the corresponding predicted future tactile feature at each time step, producing a tactile representation.
Meanwhile, the robot’s historical delta actions over the past $h$ steps are transformed into the TCP coordinate frame and concatenated with the corresponding delta gripper states to form a trajectory feature. 
Finally, the tactile representation and trajectory feature are concatenated and fed into a three-layer MLP to output a single-step refined action $a_r$.

\begin{figure}[t]
\centering
\includegraphics[width=0.8\linewidth]{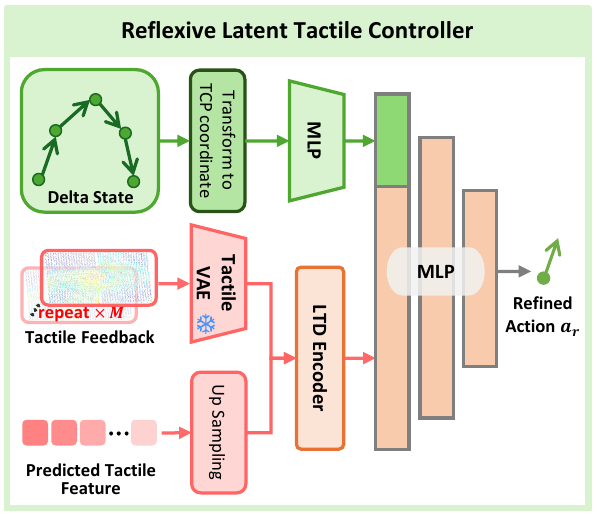}
\caption{{\textbf{Overview of the Reflexive Latent Tactile Controller (RLTC).} It takes single-frame tactile feedback, chunk-aligned predicted tactile latents, and robot/gripper delta states as input to produce high-frequency (60~Hz) refined single-step corrective actions.}}
\label{fig:controller}
\vspace{-3mm}
\end{figure}

\subsubsection{Training Loss}
To train the Reflexive Latent Tactile Controller, we jointly use abnormal recovery samples and normal-contact samples. Both are extracted from offline expert demonstrations in OmniViTac. For each trajectory, we compute the mean $\mu$ and standard deviation $\sigma$ of the tactile amplitude during contact and define its valid range as $[\mu-\sigma,\mu+\sigma]$. A recovery segment is selected if the tactile state returns to this range within 1~s after a deviation or contact loss and remains within it for at least 0.5~s. Normal-contact samples are extracted from segments that remain within the valid range.

All recovery samples from the training demonstrations are used. Normal-contact samples are drawn uniformly from the full normal-contact pool to maintain a 1:1 recovery-to-normal ratio in each minibatch. For each abnormal recovery sample, the target corrective action $\hat{a}_r$ is defined as the residual between the expert recovery action and the action generated by the slow policy, whereas the target residual for each normal-contact sample is set to zero. The controller is trained to predict the corrective action $a_r$ using a mean squared error (MSE) loss:
\begin{equation}
\mathcal{L}_{RLTC} = \| a_r - \hat{a}_r \|_2^2,
\end{equation}
which encourages the controller to output near-zero residuals during normal execution while producing high-frequency tactile-driven corrections under abnormal contact conditions.

%% file: sections/5_experiments.tex
\section{Experimental Evaluation}
We describe the experimental setup and implementation details, then evaluate OmniVTA across six contact-rich manipulation tasks under object diversity, generalization, and perturbation settings. Our ablation studies address four questions: (1) how accurately TactileVAE reconstructs tactile deformation and distinguishes contact patterns, (2) how the world model compares with alternative methods in predicting future tactile signals, (3) whether these predictions improve action planning and how best to incorporate them, and (4) how the RLTC corrects actions during execution and improves task performance.

\subsection{Experimental Setup}

\begin{table*}[t]
\centering
\caption{Objects used for training and manipulation across different robotic tasks. Each object is associated with 150 manipulation trajectories in the dataset for training world models and manipulation policies.}
\small
\renewcommand{\arraystretch}{1.2}
\begin{tabular}{@{} l l l @{}}
\toprule
\textbf{Task} & \textbf{Object Used for Training} & \textbf{Object Used for Manipulation} \\
\midrule
\textbf{Wipe} & vases in 4 different colors and shapes, plate, and whiteboard  & vases in 4 different colors and shapes \\
\textbf{Peel} & cucumber, Chinese yam, radish, carrot, and white daikon & cucumber, Chinese yam, radish \\
\textbf{Cut} & cucumber, Chinese yam, carrot, pepper, and banana & cucumber, Chinese yam, pepper, banana \\
\textbf{Assembly} & 3 colors of USB stick and 3 types of charging adapters & red USB stick and silver USB stick \\
\textbf{Grasp} & blueberry, strawberry, grape, cherry, and cherry tomato & blueberry, strawberry, grape, and cherry tomato \\
\textbf{Adjustment} & test tube, pencil, cuboid, cylinder, and triangular prism & cuboid, cylinder \\
\bottomrule
\label{tab:object}
\end{tabular}
\end{table*}

\subsubsection{Training Details}
Our OmniVTA framework is trained in four stages: training the TactileVAE, training the Visuo-Tactile World Model, training the Adaptive Visuo-Tactile Fusion Policy, and training the Reflexive Latent Tactile Controller.

\noindent\textbf{Training of TactileVAE.}
For training the TactileVAE, we use tactile data from 20\% of the manipulation trajectories, together with additional tactile interaction data collected from interactions between 10 extra objects and the tactile sensors. The resulting dataset contains approximately 1.2M tactile samples. 
Both TactileVAE and the VAE baselines are trained for 50 epochs using 8 NVIDIA A100 GPUs.
The loss weights are set to $\lambda_{\text{KL}}=1e-6$.

\noindent\textbf{Training of Visuo-Tactile World Model.}
We first construct a dataset to train the Visuo-Tactile World Model, comprising six categories of tactile interaction patterns. 
For each category, 5--6 objects are selected, and 150 manipulation trajectories per object are collected using TacUMI and the xArm7. 
The data are split into training and test sets with a 90\%/10\% ratio. 
A detailed breakdown of the dataset is provided in Table~\ref{tab:object}.

For training the Visuo-Tactile World Model, we use AdamW with a learning rate of $1\times10^{-4}$ and weight decay $0$. The per-GPU batch size is $5$, and the total number of training steps is $100{,}000$. The gradient norm threshold is set to $0.1$, with gradient clipping enabled after $20{,}000$ steps. The loss weights are set to $\lambda_{\text{dyn}}=1.0$ and $\lambda_{\text{amp}}=1.0$.

\noindent\textbf{Training of Adaptive Visuo-Tactile Fusion Policy.}
For each interaction category, OmniVTA and all policy baselines use the same task demonstrations and data splits. The Adaptive Visuo-Tactile Fusion Policy is trained for 250k steps. DP, DP+tactile, KineDex, ForceMimic, and RDP are trained for 350k steps. Dream-Tac and T-Rex follow their official training settings, input formats, and execution frequencies. We set the loss weight to $\lambda_{ct}=0.2$.

\noindent\textbf{Parameter settings.}
Following prior work, we adopt \emph{relative actions} as the action representation. We derive relative motion labels from consecutive measured end-effector poses obtained from xArm7 feedback or TacUMI pose tracking. During execution, the robot tracks the slow policy's end-effector targets after correction by the RLTC.
During training, visual observations are sampled at 15\,Hz, tactile signals at 60\,Hz, and robot proprioception at 60\,Hz. 
The policy outputs action chunks at 15\,FPS. 
For observations, OmniVTA uses the current and previous visual frames (2 frames in total), the corresponding 8 tactile frames within the same temporal window, and 2 proprioceptive observations as policy inputs.
Each action chunk predicts the next 6 actions, which are interpolated to 60\,Hz during execution.

\noindent{\textbf{Details about the model size.}}
{Table~\ref{tab:model_size} shows the parameter counts of the major components of OmniVTA, including the TactileVAE, the visuo-tactile world model, and the RLTC.}

\begin{table}[H]
    \centering
    \caption{{Model sizes of the main OmniVTA components.}}
    \label{tab:model_size}
    \footnotesize
    \setlength{\tabcolsep}{6pt}
    \renewcommand{\arraystretch}{1.10}
    \begin{tabular}{@{}lr@{}}
        \toprule
        \textbf{Component} & \textbf{Parameters} \\
        \midrule
        TactileVAE
        & 1.96\,M \\
        Visuo--Tactile World Model
        & 271.57\,M \\
        Adaptive Visuo--Tactile Fusion Policy
        & 326.89\,M \\
        RLTC
        & 0.10\,M \\
        \midrule
        \textbf{Total}
        & \textbf{600.52\,M} \\
        \bottomrule
    \end{tabular}
\end{table}

\subsubsection{Hardware Setup}

The experimental platform consists of a UFactory xArm7 robotic arm equipped with a parallel two-finger gripper and two fingertip tactile sensors. 
An Intel RealSense D435 camera mounted on the robot wrist provides RGB observations at 15\,Hz, while the tactile sensors capture tactile signals at 60\,Hz. For the TactileVAE validation experiments, three types of tactile sensors are used: GelSight Mini, Tac3D, and Xense. 
The resolutions of their reconstructed 3D displacement fields are $7\times9$, $20\times20$, and $35\times20$, respectively. 
We use Xense-Photon sensors for the main real-robot manipulation experiments.

\subsubsection{Baselines}
We select representative baseline methods to evaluate the effectiveness of the proposed TactileVAE, Visuo-Tactile World Model, and the overall OmniVTA framework. 

\noindent \textbf{TactileVAE baselines.} We compare the proposed TactileVAE with two representative tactile encoding methods: PCA-based features and a point cloud-based autoencoder~\cite{chen2024general}.

\noindent \textbf{Tactile prediction baselines.}
We compare our Visuo-Tactile World Model against three representative paradigms for multi-modal generative modeling.

\begin{itemize}
\item \textbf{Unified multimodal generation}:
UVA~\cite{li2025unified} models video frames and actions as a unified token sequence and jointly generates them using a single generative model. We extend this formulation by injecting tactile tokens.

\item \textbf{Joint trajectory diffusion}:
ForceMimic~\cite{liu2025forcemimic} and KineDex~\cite{zhang2025kinedex} concatenate interaction forces with actions and learn diffusion-based policies conditioned on visual observations. In our implementation, we replace force inputs with tactile feature embeddings extracted using a lightweight MLP autoencoder.

\item \textbf{Conditional modality prediction}:
exUMI~\cite{xu2025exumi} generates future tactile signals using a latent diffusion model (LDM) conditioned on visual observation and action. Since our method assumes only historical observations are available at inference, future action inputs are replaced with historical action sequences.
\end{itemize}

\noindent \textbf{Contact-rich policy baselines.}
We compare against several state-of-the-art imitation learning policies:

\begin{itemize}
\item \textbf{DP}~\cite{chi2025diffusion}: Diffusion Policy using wrist-view RGB observations and open-loop action chunk execution.
\item \textbf{DP+tactile}: Diffusion Policy augmented with tactile features encoded by PCA.
\item \textbf{KineDex}~\cite{zhang2025kinedex}: a diffusion policy that jointly predicts actions and forces from visual observations.
\item \textbf{ForceMimic}~\cite{liu2025forcemimic}: a diffusion policy conditioned on 3D observations that jointly predicts actions and forces.
\item \textbf{RDP}~\cite{xue2025reactive}: a reactive diffusion policy with a slow diffusion planner and a fast tactile-based reactive controller.
\item \textbf{Dream-Tac}~\cite{lou2026dreamtac}: a world-action model that jointly models actions and future visuo-tactile observations through contact-aware fusion.
\item \textbf{T-Rex}~\cite{niu2026trex}: a tactile-reactive policy with a temporal tactile VQ-VAE and a variable-rate Mixture-of-Transformers architecture.
\item \textbf{OmniVTA (ours)}: our method with a visuo-tactile world model and a dual-system manipulation policy consisting of a diffusion planner and a reflexive tactile controller.
\item \textbf{OmniVTA w/o RLTC}: our method without the reflexive controller, executing diffusion-planned action chunks in an open-loop manner.
\end{itemize}

Dream-Tac and T-Rex use official implementations and pretrained weights, with the same task demonstrations and data splits as OmniVTA; each retains its official training settings, input format, and execution frequency.

\begin{figure}[ht]
\centering
\includegraphics[width=0.75\linewidth]{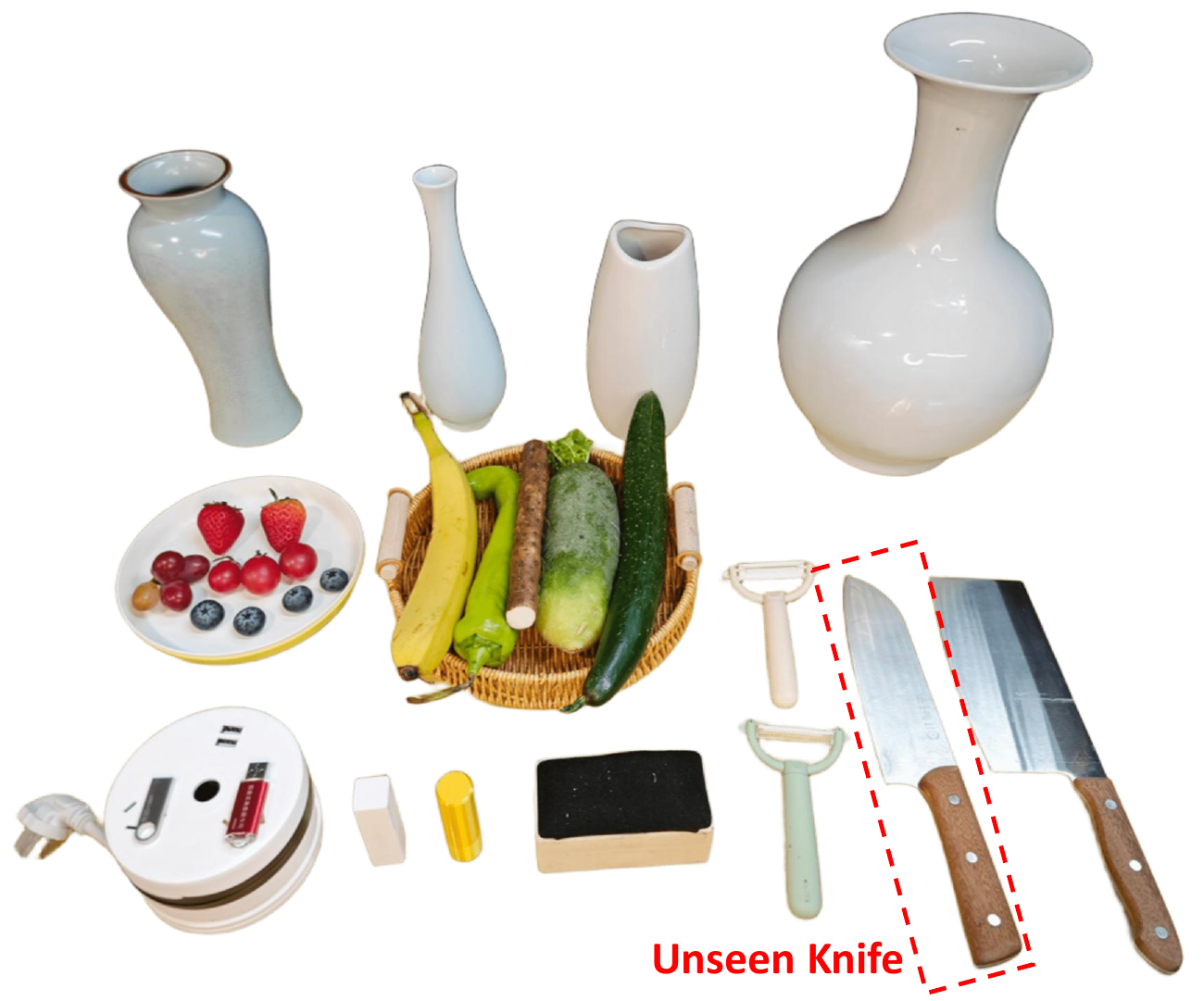}
\caption{\textbf{Objects used in the manipulation tasks.}}
\label{fig:object}
\vspace{-3mm}
\end{figure}

\subsection{Overall Performance}

We evaluate the real-world manipulation performance of OmniVTA from three perspectives: object diversity, generalization, and perturbation robustness.

\noindent \textbf{Object diversity.}
For each category, we evaluate the policy on 2--4 different objects, as summarized in Table~\ref{tab:object} and Fig.~\ref{fig:object}. 
Before each trial, the object is placed at a random position and fixed during manipulation. 
For each object, we perform 10 trials and report the average success rate.

\noindent \textbf{Generalization.}
We evaluate two types of generalization.

(1) \emph{Position generalization.}
For the wipe, peel, assembly, and adjustment tasks, objects are placed at two additional unseen heights during evaluation. 
For each object and each height, we conduct 5 trials and report the average success rate.

(2) \emph{Tool generalization.}
For the cut task, we replace the training knife with a previously unseen knife (shown in Fig.~\ref{fig:object}). 
Each object is tested 10 times using the new tool.

\noindent \textbf{Perturbation robustness.}
For the wipe, peel, cut, and assembly tasks, we introduce controlled perturbations during execution to disrupt the current contact state. Specifically, the target object is suddenly displaced upward or downward along the vertical direction during the interaction phase. This perturbation breaks the existing contact state and requires the policy to re-establish stable contact. For wiping, peeling, and cutting, we test one representative object per task over 10 trials. For assembly, we test two USB plugs (red and silver) over 10 trials each, for a total of 20 trials.

\noindent \textbf{Evaluation metric.}
For wiping, peeling, and cutting, we report the mean fraction of the target length processed across trials. For assembly, grasping, and adjustment, we report success rates. Success requires full insertion, a secure grasp, or an orientation change exceeding $60^\circ$, respectively. All tasks require the tactile sensors to remain undamaged.

\begin{figure*}[t]
\centering
\includegraphics[width=0.9\linewidth]{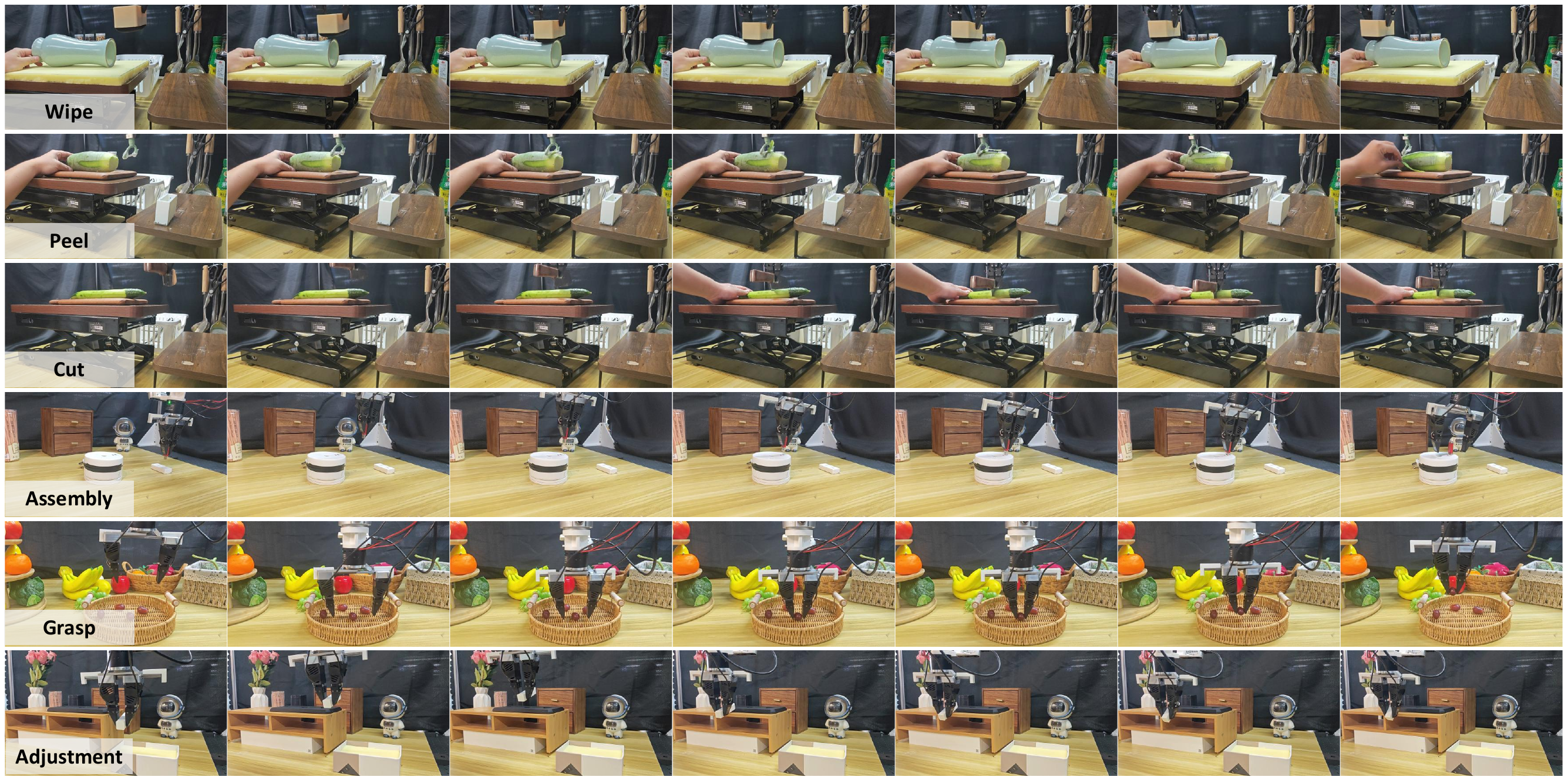}
\caption{\textbf{Real-robot manipulation results.} We visualize the manipulation process across six task categories.}
\label{fig:manipulation}
\end{figure*}

\noindent \textbf{Results.}
The overall real-robot manipulation results are visualized in Fig.~\ref{fig:manipulation}, and the task scores are reported in Table~\ref{tab:full_comparison_final_scaled}. OmniVTA achieves the highest score outright in 11 of the 15 conditions and ties with RDP in assembly O and G; T-Rex leads in peeling G and grasping O.
The mean and maximum tangential deformations are 0.35 and 0.72 for OmniVTA, compared with 0.56 and 1.1 for RDP. At unseen heights, OmniVTA without RLTC exceeds RDP only on wiping (0.40 vs.\ 0.38), while the full model exceeds RDP on wiping, peeling, and adjustment and ties on assembly. With the unseen knife, OmniVTA scores 0.83, compared with 0.60 for T-Rex and 0.35 for Dream-Tac. Under perturbations, RLTC improves performance in all four tasks, and the full model achieves the highest scores.

\input{tables/result}

\vspace{-3mm}
\subsection{Component Analysis}
\subsubsection{TactileVAE}
We evaluate TactileVAE from two aspects: reconstruction accuracy and latent representation quality.

\noindent \textbf{Reconstruction evaluation.}
We compare TactileVAE with baseline methods across six categories of tactile interaction patterns. We measure reconstruction error using the $L_2$ distance over the entire 3D deformation field and evaluate directional consistency using cosine similarity over regions with nonzero deformation. As shown in Table~\ref{tab:vae}, TactileVAE achieves the lowest $L_2$ error in all six categories and the highest cosine similarity in five. For grasping, TactileVAE achieves a lower $L_2$ error than PointNet-AE, while PointNet-AE achieves higher cosine similarity (0.750 vs.\ 0.720).

\input{tables/vae_exp}

\begin{figure}[ht]
\centering
\includegraphics[width=\linewidth]{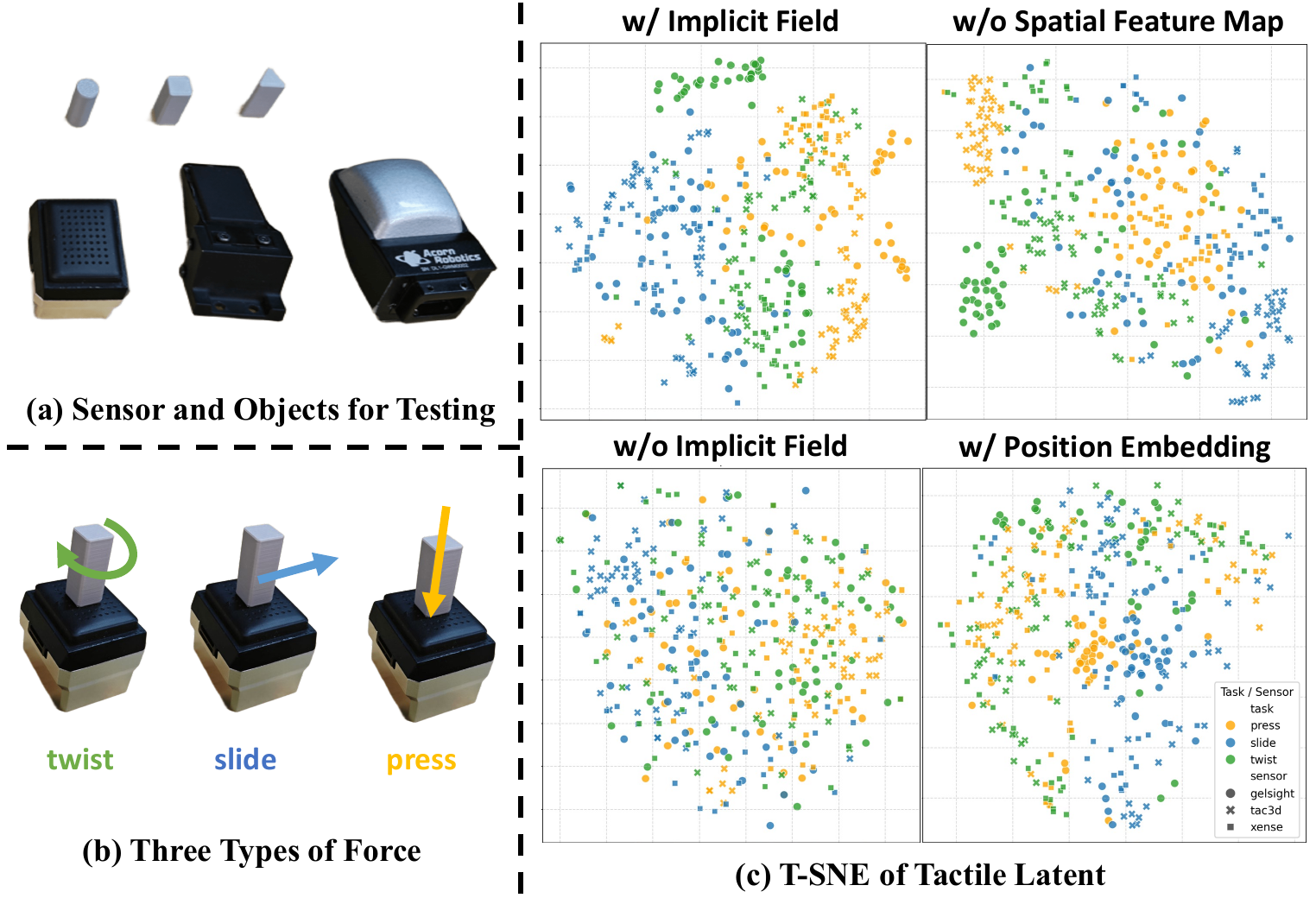}
\caption{\textbf{Ablation study of TactileVAE.}\textbf{(a)} shows the sensors and objects for testing. \textbf{(b)} represents the three types of force loaded on the sensors. \textbf{(c)} visualizes the t-SNE of the tactile latent feature maps extracted by different models.}
\label{fig:vae_tsne}
\vspace{-3mm}
\end{figure}

\noindent \textbf{Latent representation analysis.}
To further evaluate the learned tactile representations, we conduct experiments on three unseen objects under three types of force patterns (press, twist, and slide) using three tactile sensors, two of which are unseen during training, as shown in Fig.~\ref{fig:vae_tsne}(a) and (b). We record the tactile signals and extract their features using several variants of TactileVAE, including:
\begin{itemize}
    \item \textbf{w/o implicit decoder:} decodes the latent feature map without query-based sampling and query points input.
    \item \textbf{w/ position embedding:} incorporates positional encoding during decoding.
    \item \textbf{w/o spatial feature map:} represents the latent as a single token and decodes it with an implicit decoder.
    \item \textbf{w/ implicit decoder:} decodes the latent feature map via an implicit decoder.
\end{itemize}
All models are trained using only 5\% of the tactile dataset.

For visualization, the spatio-temporal tactile features are aggregated via max pooling to obtain sequence-level embeddings, which are then visualized using t-SNE. As shown in Fig.~\ref{fig:vae_tsne}(c), the learned representations exhibit clear clustering across different force patterns, even under cross-sensor settings. In contrast, marker-based encoding and variants without positional information fail to produce well-separated clusters, highlighting the importance of implicit neural representations for learning discriminative tactile features.

Local feature maps reduce reconstruction error relative to a pooled token on all three sensors (Table~\ref{tab:vae_ablation}). Reconstructing local 3D deformation from unlabeled frames yields a shared representation for tactile prediction and contact feedback. On an RTX 4090D, single-finger encoding takes 1.82~ms for TactileVAE and 11.09~ms for AnyTouch~2~\cite{anytouch2} with their native inputs. AnyTouch~2 aligns touch with vision and language, while FG-CLTP~\cite{fgcltp} aligns tactile point clouds with descriptions of contact geometry and force. TactileVAE supports fast feedback but cannot directly infer object category or function from touch.

\input{tables/vae_ablation}

\subsubsection{Visuo-Tactile World Model}

\begin{figure*}[ht]
\centering
\includegraphics[width=\linewidth]{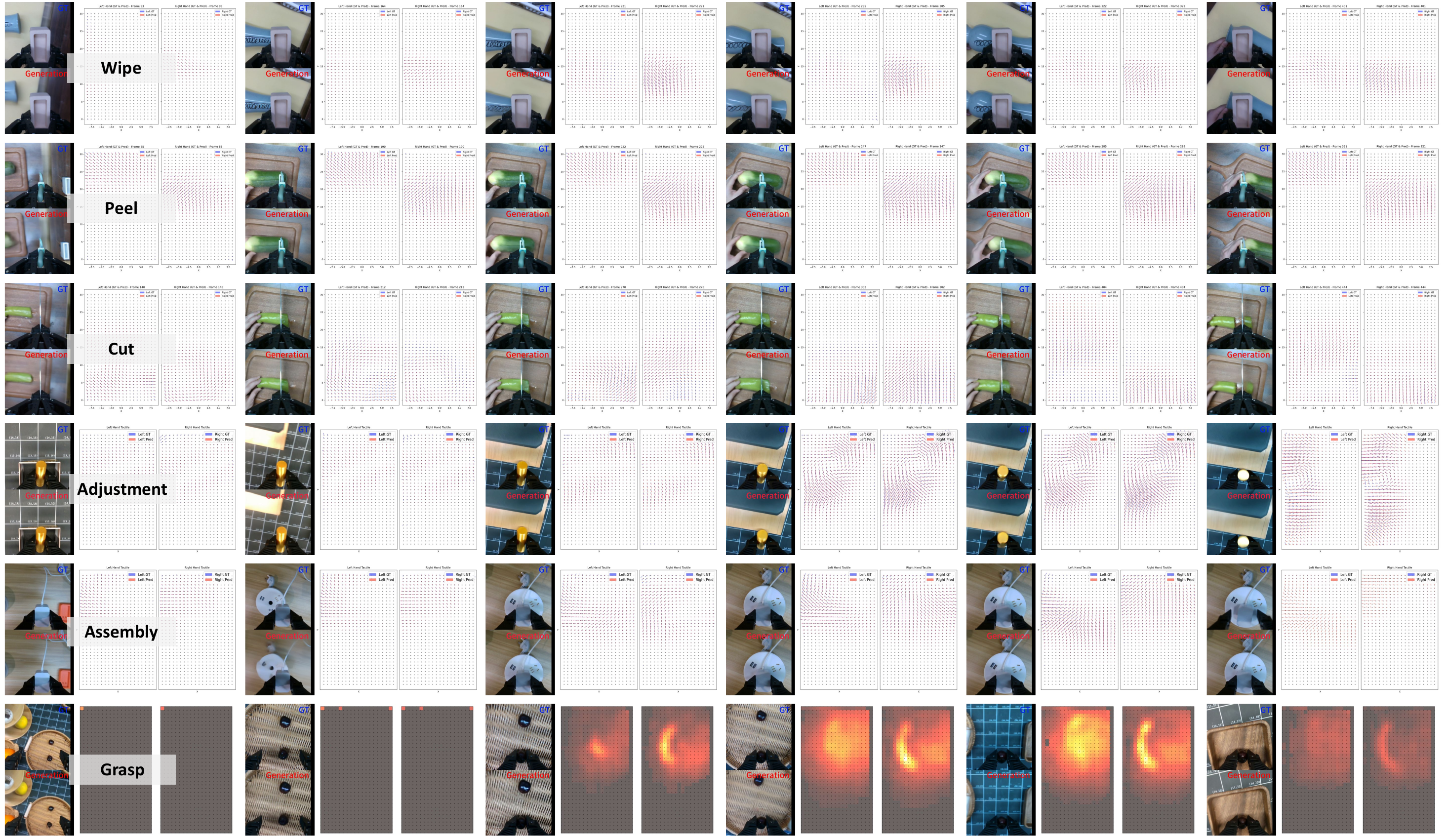}
\caption{\textbf{Visuo-tactile generation visualization across six task categories.} Red arrows indicate predicted tangential deformation, while blue arrows denote ground truth. Normal deformation in the grasp task is visualized using heatmaps. }
\label{fig:wm}
\end{figure*}

\input{tables/wm_exp}
\input{tables/ablation_wm}

In this section, we evaluate the proposed Visuo-Tactile World Model in terms of prediction accuracy, module effectiveness, and robustness to observation perturbations.
We measure tactile prediction accuracy using the $L_2$ distance and cosine similarity between the predicted and ground-truth tactile signals, following the evaluation
protocol of TactileVAE.
To better illustrate prediction performance over different horizons, we report both the $L_2$ distance and cosine similarity at the 2nd, 4th, and 6th prediction latent frames,
corresponding to the 8th, 16th, and 24th decoded tactile frames, respectively, as well as the metrics averaged over all 24 predicted frames.


\noindent \textbf{Prediction accuracy.}
We first compare our method with four baselines from three representative paradigms: unified multimodal generation, joint trajectory diffusion, and conditional tactile prediction. 
As shown in Table~\ref{tab:wm_exp} and Fig.~\ref{fig:wm}, our model consistently outperforms all baselines in both short-term and long-term tactile prediction. These results demonstrate that our method effectively models the cross-modal dynamics between vision and tactile signals, enabling accurate and consistent generation across both modalities.

\noindent \textbf{Ablation study.}
We further analyze the contribution of several key components in the proposed model.

(1) \emph{Joint visual–tactile generation.} 
Table~\ref{tab:ablation_wm} shows that jointly generating future visual features together with tactile features improves tactile prediction accuracy. 
This indicates that visual prediction provides complementary global dynamic cues that help the model better capture interaction dynamics.

(2) \emph{Action representation.} 
We compare four representations of executed gripper motion: absolute 3D, relative 3D, 2D image projections, and 2D projections augmented with depth. The last two use the same historical gripper positions expressed in the current camera frame, with the depth variant also retaining each position's optical-axis coordinate. At unseen object positions, the 2D projection yields lower $L_2$ error than either 3D representation (Table~\ref{tab:ablation_wm}). Adding depth further reduces $L_2$ from 0.042 to 0.037 at unseen positions and from 0.035 to 0.033 at seen positions. Depth thus complements image-plane motion in these tests and can distinguish motions along the optical axis that share the same image projection.

(3) \emph{Dynamic weighting.}
The results in Table~\ref{tab:ablation_wm} show that the proposed dynamic weighting strategy improves tactile prediction performance. 
This weighting mechanism explicitly emphasizes regions with rapid dynamics and strong contact responses during optimization, enabling the model to better capture high-frequency tactile patterns, local contact variations, and fine-grained temporal structures.

\begin{figure*}[ht]
\centering
\includegraphics[width=0.9\linewidth]{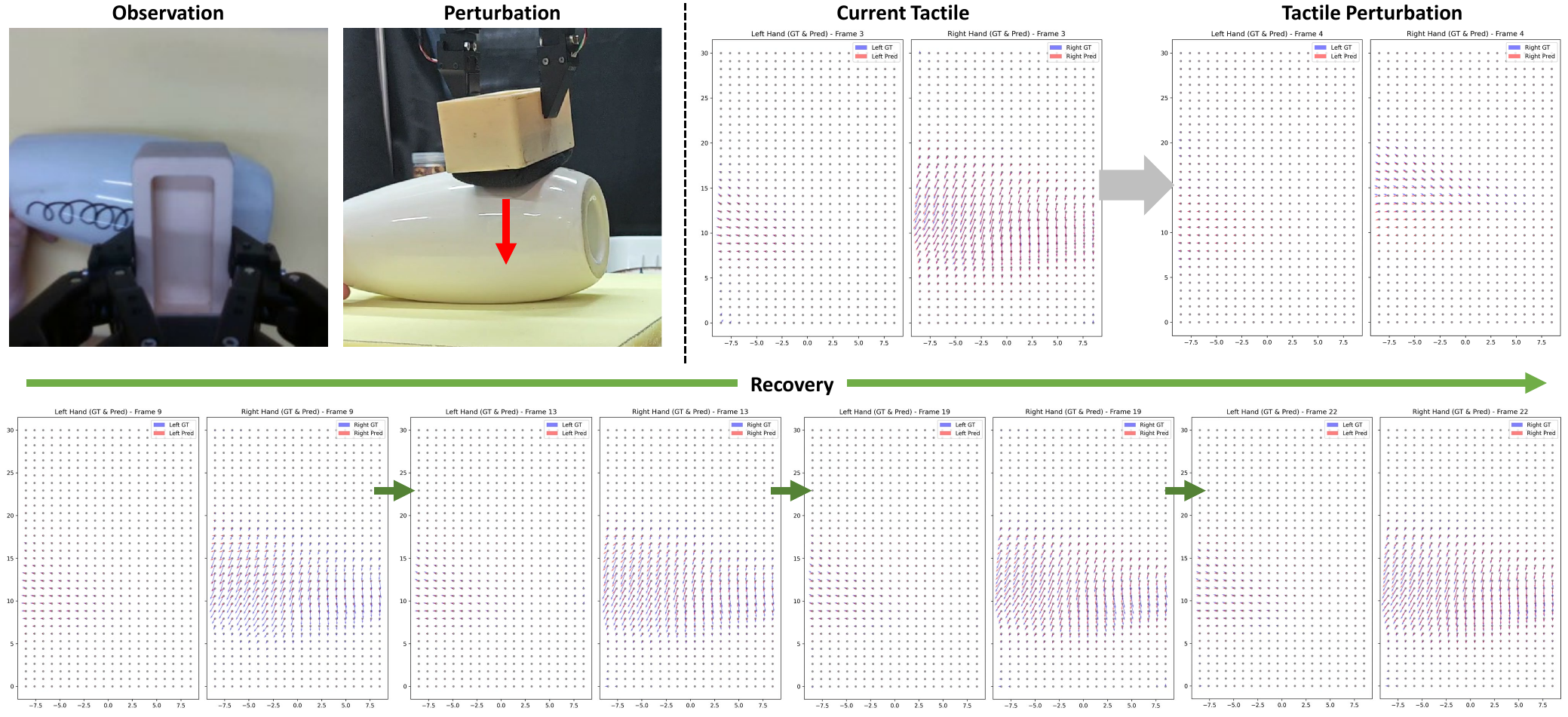}
\caption{\textbf{Perturbation and recovery.} The first row shows a contact break due to downward perturbation, with tactile signals transitioning to a no-contact state. The second row shows recovery to the contact state within one generation chunk. In tactile visualization, red arrows denote tactile prediction while blue arrows denote ground truth.}
\label{fig:wm_disturb}
\vspace{-3mm}
\end{figure*}

\noindent \textbf{Perturbation robustness.}
Finally, we evaluate the robustness of the Visuo-Tactile World Model in the wipe-vase task. 
After contact occurs, the visual and tactile observations are replaced with non-contact observations corresponding to the object being moved downward. 
As shown in Fig.~\ref{fig:wm_disturb}, the model exhibits a certain level of recovery capability under observation perturbations.

\subsubsection{Adaptive Visuo-Tactile Fusion Policy}

In this section, we analyze the key components of the proposed Adaptive Visuo-Tactile Fusion Policy through a series of experiments and address the following questions:
(1) Does predicted future tactile information improve action planning?
(2) Is the proposed Latent Tactile Differential (LTD) Encoder more effective than simple feature concatenation?
(3) Is the gating mechanism necessary for adaptive visuo-tactile fusion, and do modality weights correlate with contact states?
(4) Does the Visuo-Tactile Diffusion Policy depend on the accuracy of tactile prediction?
(5) Do generated visual features further improve policy performance?
(6) What role does the Reflexive Latent Tactile Controller play during execution?

\noindent \textbf{Effect of predicted tactile signals.}
We first investigate whether predicted tactile information improves action planning. Table~\ref{tab:ablation_components_scaled} compares two settings: (1) using only the current tactile observation, with gating disabled (row~1), and (2) concatenating current and predicted tactile features at horizons of two, four, or six steps, with gating enabled (rows~2--4). The concatenated tactile features are projected to the same dimension as the current tactile representation through a linear layer. Within setting~(2), extending the horizon from two to four to six steps raises the mean task score from 0.33 to 0.38 to 0.40. We use six steps because this horizon gives the highest wiping score and matches the best peeling score among the three horizons.

\input{tables/ablation_policy}
\input{tables/policy_detail}

\noindent \textbf{Comparison of tactile representations.}
We compare two ways of representing current and predicted touch: (1) direct concatenation of the current and predicted tactile features and (2) differential encoding of the same features with the proposed Latent Tactile Differential (LTD) Encoder. Both settings use a six-step prediction horizon, the same tactile feature dimension, and gated visuo-tactile fusion. Under these matched settings, LTD improves the wiping score from 0.50 to 0.66 and the peeling score from 0.30 to 0.40, showing that explicitly encoding tactile changes benefits both tasks (Table~\ref{tab:ablation_components_scaled}, rows~4 and~7).

\begin{figure*}[t]
\centering
\includegraphics[width=0.95\linewidth]{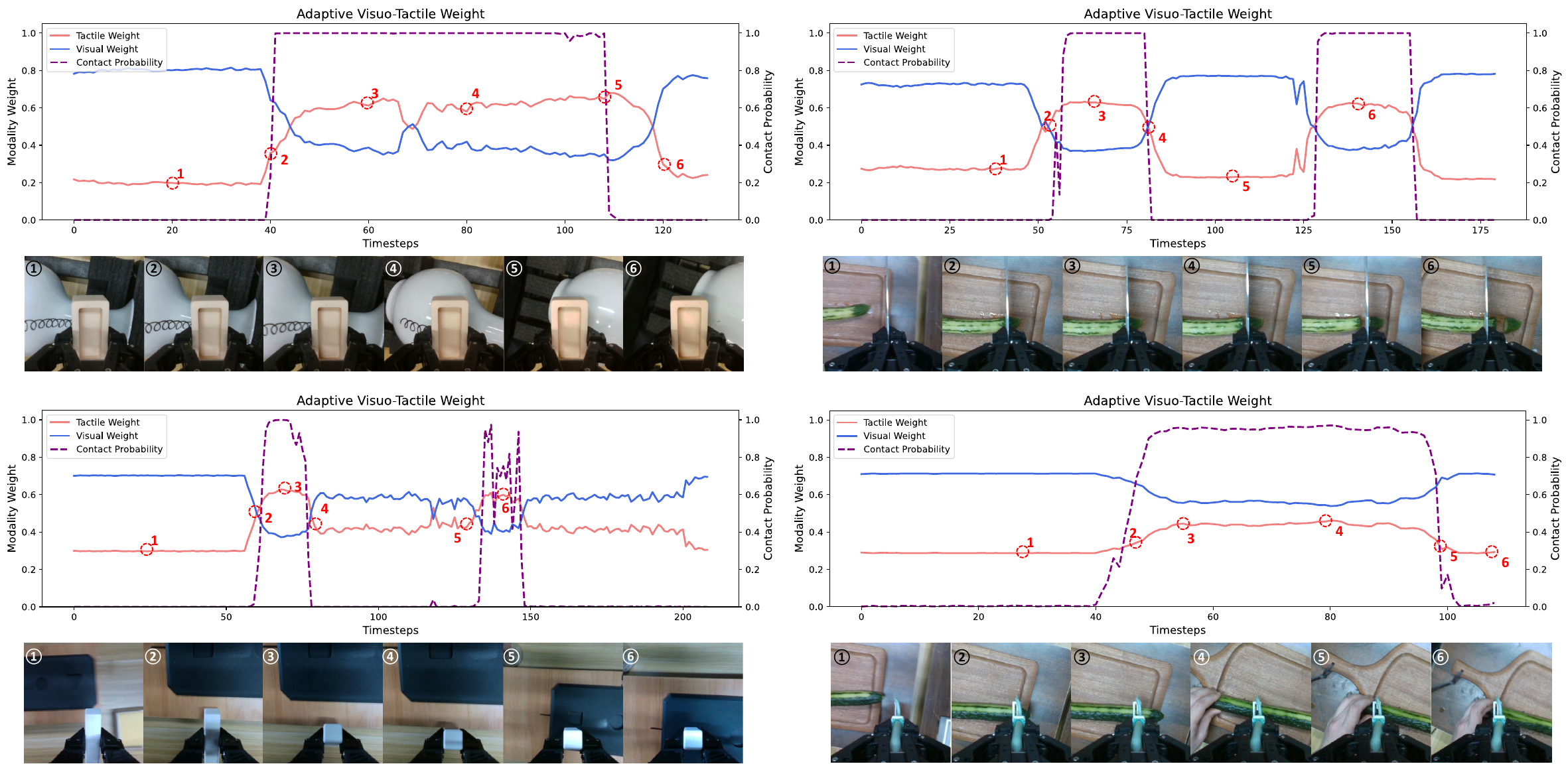}
\caption{\textbf{Gating Fusion Mechanism.} We visualize the temporal evolution of the predicted contact probability (purple), together with the corresponding visual (blue) and tactile (red) weights over the course of task execution.}
\label{fig:gate_weight}
\vspace{-3mm}
\end{figure*}

\begin{figure*}[ht]
\centering
\includegraphics[width=0.8\linewidth]{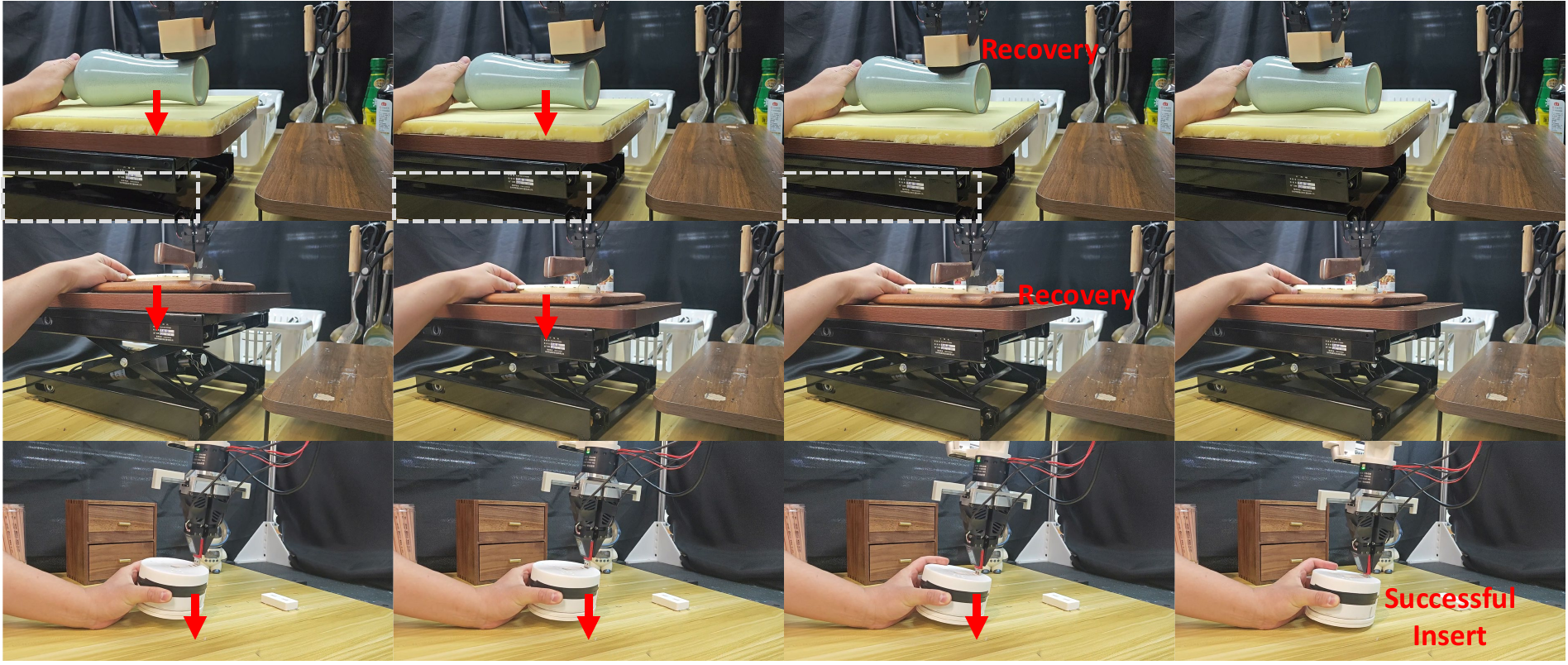}
\caption{\textbf{Perturbation experiments.} We break contact by abruptly lowering the object and demonstrate how the policy rapidly restores contact via the RLTC.}
\label{fig:mp_disturb}
\vspace{-3mm}
\end{figure*}

\begin{figure}[t]
\centering
\setlength{\abovecaptionskip}{1pt}
\includegraphics[width=\linewidth]{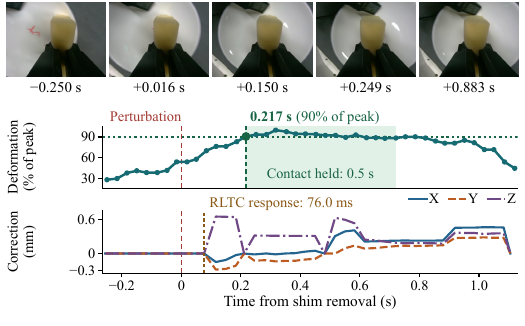}
\caption{\textbf{RLTC response during plate wiping.} From top to bottom: right-wrist RGB frames, tactile deformation normalized to the recovery peak, and right-arm corrective actions along the three axes. Shim removal defines perturbation onset ($t=0$). Shading marks 0.5~s of continuous contact after deformation first reaches 90\% of the peak.}
\label{fig:time_ctrl}
\end{figure}

\noindent \textbf{Necessity of the gating fusion mechanism.}
Next, we evaluate the effectiveness of the gating mechanism for visuo-tactile fusion by comparing two input configurations: (1) direct concatenation of visual features and LTD-encoded tactile features, and (2) adaptive fusion using the proposed gating mechanism. With the tactile representation fixed, gated fusion raises the mean task score from 46.5\% to 53.0\%, a gain of 6.5 percentage points (Table~\ref{tab:ablation_components_scaled}, rows~6 and~7). Figure~\ref{fig:gate_weight} illustrates the learned weighting: the tactile weight stays near zero during non-contact intervals and rises with the predicted contact probability. Thus, the gate gives touch greater weight during contact and improves task performance over direct concatenation.

\begin{figure}[ht]
\centering
\includegraphics[width=\linewidth]{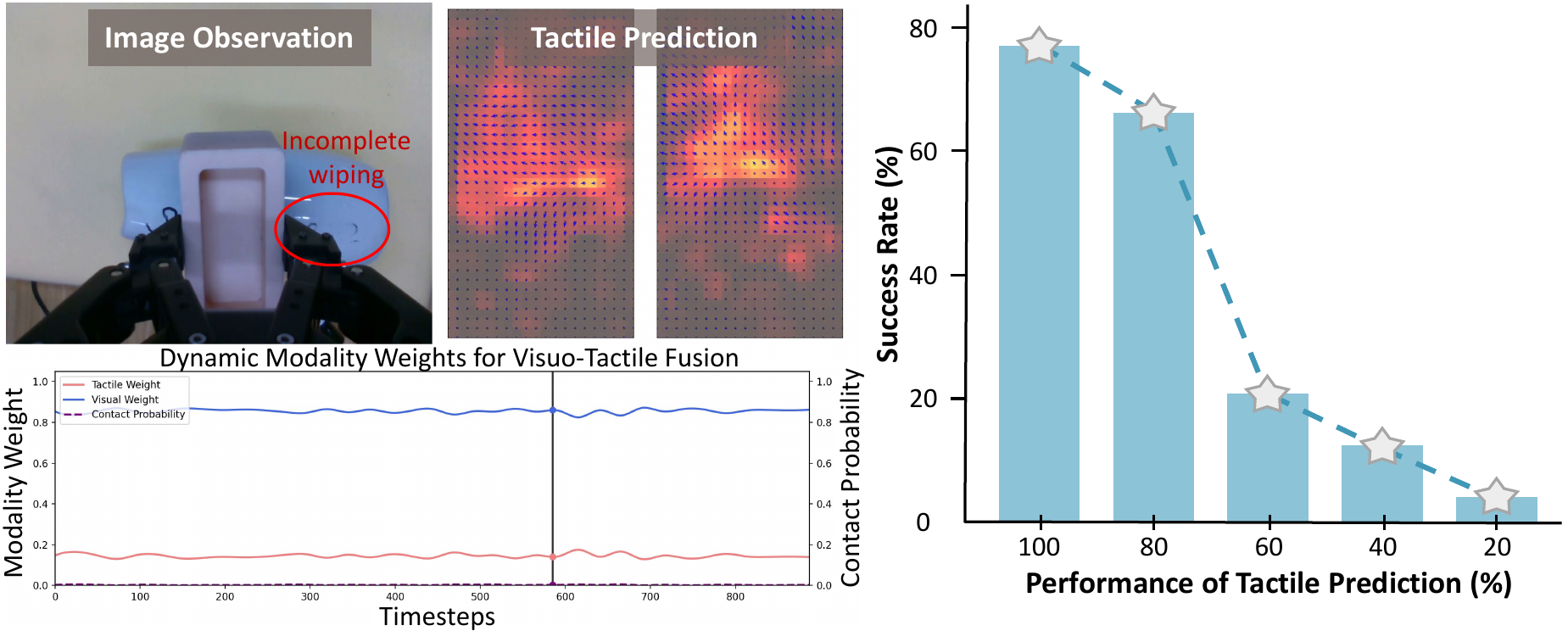}
\caption{\textbf{Effect of tactile prediction accuracy.} \textbf{Left:} At 60\% prediction performance, the model fails to estimate future contact probability, resulting in improper modality weighting. \textbf{Right:} Lower prediction performance leads to a significant drop in success rate.}
\label{fig:prediction}
\vspace{-5mm}
\end{figure}

\noindent \textbf{Effect of tactile prediction.}
We evaluate policy sensitivity to tactile prediction quality using world-model checkpoints at 80\%, 60\%, 40\%, and 20\% of the best checkpoint's prediction performance. As prediction quality declines, the policy tracks contact transitions less accurately, adjusts its modality weights less reliably, and achieves lower task scores (Fig.~\ref{fig:prediction}). The contact-state-only configuration also scores below the default policy with LTD and gating (0.37 vs.\ 0.53 on average in Table~\ref{tab:ablation_components_scaled}). These results show that the policy benefits from accurate predictions of how contact deformation evolves.

\noindent \textbf{Effect of generated visual features.}
Adding predicted visual features raises the wiping score from 0.66 to 0.70 but lowers the peeling score from 0.40 to 0.38 (Table~\ref{tab:ablation_components_scaled}, rows~7 and~8). The mean score increases by only 0.01, while the additional generation branch reduces inference frequency (Table~\ref{tab:policy_time}). We therefore use current visual observations together with predicted tactile features to retain the higher control rate.

\noindent \textbf{Effectiveness of the RLTC.}
Enabling RLTC improves task scores in all four perturbation tests (Table~\ref{tab:full_comparison_final_scaled}). In the plate-wiping trial (Fig.~\ref{fig:time_ctrl}), the RLTC correction began to change 76.0~ms after shim removal. Tactile deformation reached 90\% of its recovery peak at 0.217~s, and contact persisted for at least the next 0.5~s. The robot then completed the wiping task.

%% file: tables/result.tex
\begin{table*}[t]
\centering
\caption{Performance comparison of OmniVTA and baseline methods across six tasks under three evaluation settings: object diversity (O), generalization (G), and perturbation robustness (P). Scores denote mean completion fractions for wiping, peeling, and cutting, and success rates for assembly, grasping, and adjustment.}
\label{tab:full_comparison_final_scaled}
\small
\setlength{\tabcolsep}{2.5pt}
\renewcommand{\arraystretch}{1.3}

\begin{tabularx}{\textwidth}{@{} l *{15}{>{\centering\arraybackslash}X} @{}}
\toprule
\textbf{Task} & \multicolumn{3}{c}{\textbf{Wipe}} & \multicolumn{3}{c}{\textbf{Peel}} & \multicolumn{3}{c}{\textbf{Cut}} & \multicolumn{3}{c}{\textbf{Assembly}} & \multicolumn{1}{c}{\textbf{Grasp}} & \multicolumn{2}{c}{\textbf{Adjustment}} \\
\cmidrule(lr){2-4} \cmidrule(lr){5-7} \cmidrule(lr){8-10} \cmidrule(lr){11-13} \cmidrule(lr){14-14} \cmidrule(lr){15-16}
\makecell[l]{\textbf{Method /}\textbf{ Eval. setting}} & \textbf{O} & \makecell[c]{{\textbf{G}}} & \textbf{P} & \textbf{O} & \makecell[c]{\textbf{G}} & \textbf{P} & \textbf{O} & \makecell[c]{{\textbf{G}}} & \textbf{P} & \textbf{O} & \makecell[c]{{\textbf{G}}} & \textbf{P} & \textbf{O} & \textbf{O} & \makecell[c]{\textbf{G}} \\
\midrule
DP               & 0.12 & 0.05 & 0 & 0.06 & 0 & 0 & 0.28 & 0.10 & 0 & 0.10 & 0 & 0.05 & 0.20 & 0 & 0 \\
DP+tactile       & 0.36 & 0.28 & 0 & 0.32 & 0.20 & 0.08 & 0.33 & 0.15 & 0.13 & 0.30 & 0.10 & 0.10 & 0.48 & 0.25 & 0.15 \\
KineDex          & 0.40 & 0.25 & 0 & 0.24 & 0.13 & 0.05 & 0.38 & 0.30 & 0.20 & 0.30 & 0.15 & 0.15 & 0.65 & 0.30 & 0.20 \\
ForceMimic       & 0.33 & 0.20 & 0 & 0.27 & 0.18 & 0 & 0.50 & 0.25 & 0.05 & 0.35 & 0.15 & 0.10 & 0.60 & 0.10 & 0 \\
RDP              & 0.50 & 0.38 & 0.42 & 0.48 & 0.36 & 0.45 & 0.65 & 0.50 & 0.43 & \textbf{0.60} & \textbf{0.50} & 0.35 & 0.88 & 0.50 & 0.50 \\
Dream-Tac & 0.30 & 0.22 & 0.05 & 0.24 & 0.10 & 0 & 0.50 & 0.35 & 0.10 & 0.50 & 0.40 & 0 & 0.60 & 0.50 & 0.20 \\
T-Rex & 0.70 & 0.44 & 0.56 & 0.50 & \textbf{0.50} & 0.42 & 0.60 & 0.60 & 0.36 & 0.45 & 0.40 & 0.25 & \textbf{0.95} & 0.60 & 0.50 \\
OmniVTA w/o RLTC & 0.66 & 0.40 & 0.25 & 0.40 & 0.30 & 0.20 & 0.50 & 0.50 & 0.20 & 0.40 & 0.35 & 0.20 & 0.70 & 0.40 & 0.30 \\
\textbf{OmniVTA} & \textbf{0.80} & \textbf{0.58} & \textbf{0.60} & \textbf{0.55} & 0.48 & \textbf{0.63} & \textbf{0.85} & \textbf{0.83} & \textbf{0.60} & \textbf{0.60} & \textbf{0.50} & \textbf{0.40} & 0.90 & \textbf{0.65} & \textbf{0.65} \\
\bottomrule
\end{tabularx}
\vspace{-3mm}
\end{table*}

%% file: tables/vae_exp.tex
\begin{table*}[t]
\centering
\caption{Performance comparison of different methods across robotic tasks using L2 and cosine similarity metrics. Lower L2 ($\downarrow$) and higher cos ($\uparrow$) indicate better performance.}
\label{tab:vae}
\small 
\renewcommand{\arraystretch}{1.2}

\begin{tabular*}{\textwidth}{@{\extracolsep{\fill}} l ccccc ccccc cc}
\toprule
\textbf{Method} & \multicolumn{2}{c}{\textbf{Wipe}} & \multicolumn{2}{c}{\textbf{Peel}} & \multicolumn{2}{c}{\textbf{Cut}} & \multicolumn{2}{c}{\textbf{Assembly}} & \multicolumn{2}{c}{\textbf{Grasp}} & \multicolumn{2}{c}{\textbf{Adjustment}} \\
\cmidrule(lr){2-3} \cmidrule(lr){4-5} \cmidrule(lr){6-7} \cmidrule(lr){8-9} \cmidrule(lr){10-11} \cmidrule(lr){12-13}
& {L2 $\downarrow$} & {cos $\uparrow$} & {L2 $\downarrow$} & {cos $\uparrow$} & {L2 $\downarrow$} & {cos $\uparrow$} & {L2 $\downarrow$} & {cos $\uparrow$} & {L2 $\downarrow$} & {cos $\uparrow$} & {L2 $\downarrow$} & {cos $\uparrow$} \\
\midrule
PCA         & 0.091 & 0.810 & 0.085 & 0.430 & 0.109 & 0.400 & 0.071 & 0.720 & 0.036 & 0.600 & 0.069 & 0.560 \\
PointNet-AE & 0.059 & 0.910 & 0.067 & 0.850 & 0.062 & 0.840 & 0.058 & 0.900 & 0.028 & \textbf{0.750} & 0.047 & 0.760 \\
\textbf{Ours} & \textbf{0.038} & \textbf{0.930} & \textbf{0.033} & \textbf{0.880} & \textbf{0.031} & \textbf{0.940} & \textbf{0.022} & \textbf{0.910} & \textbf{0.011} & 0.720 & \textbf{0.017} & \textbf{0.850} \\
\bottomrule
\end{tabular*}
\end{table*}

%% file: tables/vae_ablation.tex
\begin{table}[t]
\centering
\caption{Ablation study on the design of TactileVAE. We evaluate reconstruction performance using the $L_2$ metric.}
\label{tab:vae_ablation}
\small
\renewcommand{\arraystretch}{1.2} 
\setlength{\tabcolsep}{2pt} 

\begin{tabular}{l S[table-format=1.3] S[table-format=1.3] S[table-format=1.3]}
\toprule
\textbf{Method} & {\textbf{GelSight-Mini}} & {\textbf{Tac3D}} & {\textbf{Xense-Photon}} \\
\midrule
w/o implicit decoder & 0.126 & 0.098 & 0.038 \\
w/ position embed. & 0.102 & 0.085 & 0.035 \\
w/o spatial feature map & 0.107 & 0.084 & 0.071 \\
w/ implicit decoder & \textbf{0.047} & \textbf{0.058} & \textbf{0.034} \\
\bottomrule
\end{tabular}
\end{table}

%% file: tables/wm_exp.tex

\begin{table*}[t]
\centering
\scriptsize
\setlength{\tabcolsep}{2pt}
\renewcommand{\arraystretch}{1.15}
\caption{Prediction performance. We report the $L_2$ distance and cosine similarity at the 2nd, 4th, and 6th latent prediction steps (denoted as $L2_1, L2_2, L2_3$ and $C_1, C_2, C_3$), corresponding to the 8th, 16th, and 24th decoded tactile frames, respectively. We also report the average performance over all 24 predicted frames ($L2_{\text{avg}}$, $C_{\text{avg}}$). Top: Wipe/Peel/Cut. Bottom: Adjustment/Assembly/Grasp. ``--" denotes not applicable.}
\label{tab:wm_exp}
\resizebox{\textwidth}{!}{%
\begin{tabular}{l|cccccccc|cccccccc|cccccccc}
\hline
\multicolumn{1}{c|}{} &
\multicolumn{8}{c|}{\textbf{Wipe}} &
\multicolumn{8}{c|}{\textbf{Peel}} &
\multicolumn{8}{c}{\textbf{Cut}} \\
\hline
\textbf{Method} &
$L2_1$ & $L2_2$ & $L2_3$ & $L2_{\text{avg}}$ & $C_1$ & $C_2$ & $C_3$ & $C_{\text{avg}}$ &
$L2_1$ & $L2_2$ & $L2_3$ & $L2_{\text{avg}}$ & $C_1$ & $C_2$ & $C_3$ & $C_{\text{avg}}$ &
$L2_1$ & $L2_2$ & $L2_3$ & $L2_{\text{avg}}$ & $C_1$ & $C_2$ & $C_3$ & $C_{\text{avg}}$ \\
\hline
\textbf{UVA} &
0.084 & 0.091 & 0.088 & 0.088 & 0.68 & 0.65 & 0.64 & 0.66 &
0.090 & 0.097 & 0.102 & 0.097 & 0.62 & 0.60 & 0.60 & 0.61 &
0.075 & 0.076 & 0.078 & 0.077 & 0.72 & 0.71 & 0.71 & 0.71 \\

\textbf{exUMI} &
0.097 & 0.098 & 0.103 & 0.101 & 0.59 & 0.58 & 0.58 & 0.58 &
0.093 & 0.099 & 0.102 & 0.097 & 0.63 & 0.62 & 0.62 & 0.62 &
0.083 & 0.086 & 0.088 & 0.086 & 0.73 & 0.72 & 0.71 & 0.72 \\

\textbf{Kinedex} &
0.081 & 0.084 & 0.085 & 0.082 & 0.82 & 0.80 & 0.81 & 0.81 &
0.064 & 0.066 & 0.068 & 0.066 & 0.77 & 0.79 & 0.80 & 0.79 &
0.091 & 0.100 & 0.104 & 0.096 & 0.74 & 0.73 & 0.72 & 0.73 \\

\textbf{Forcemimic} &
0.089 & 0.091 & 0.094 & 0.091 & 0.70 & 0.68 & 0.68 & 0.68 &
0.077 & 0.077 & 0.078 & 0.077 & 0.76 & 0.76 & 0.73 & 0.76 &
0.089 & 0.091 & 0.091 & 0.090 & 0.71 & 0.71 & 0.72 & 0.71 \\

\textbf{Ours} &
\textbf{0.052} & \textbf{0.062} & \textbf{0.072} & \textbf{0.059} & \textbf{0.93} & \textbf{0.92} & \textbf{0.91} & \textbf{0.93} &
\textbf{0.033} & \textbf{0.037} & \textbf{0.041} & \textbf{0.036} & \textbf{0.88} & \textbf{0.88} & \textbf{0.87} & \textbf{0.87} &
\textbf{0.048} & \textbf{0.052} & \textbf{0.057} & \textbf{0.050} & \textbf{0.89} & \textbf{0.89} & \textbf{0.88} & \textbf{0.88} \\
\hline
\hline
\multicolumn{1}{c|}{} &
\multicolumn{8}{c|}{\textbf{Adjustment}} &
\multicolumn{8}{c|}{\textbf{Assembly}} &
\multicolumn{8}{c}{\textbf{Grasp}} \\
\hline
\textbf{Method} &
$L2_1$ & $L2_2$ & $L2_3$ & $L2_{\text{avg}}$ & $C_1$ & $C_2$ & $C_3$ & $C_{\text{avg}}$ &
$L2_1$ & $L2_2$ & $L2_3$ & $L2_{\text{avg}}$ & $C_1$ & $C_2$ & $C_3$ & $C_{\text{avg}}$ &
$L2_1$ & $L2_2$ & $L2_3$ & $L2_{\text{avg}}$ & $C_1$ & $C_2$ & $C_3$ & $C_{\text{avg}}$ \\
\hline
\textbf{UVA} &
0.080 & 0.085 & 0.084 & 0.083 & 0.73 & 0.68 & 0.67 & 0.69 &
0.072 & 0.074 & 0.078 & 0.074 & 0.74 & 0.69 & 0.68 & 0.68 &
0.079 & 0.080 & 0.085 & 0.080 & 0.67 & 0.64 & 0.64 & 0.65 \\

\textbf{exUMI} &
0.080 & 0.087 & 0.083 & 0.083 & 0.64 & 0.60 & 0.60 & 0.61 &
0.096 & 0.096 & 0.103 & 0.096 & 0.59 & 0.53 & 0.52 & 0.53 &
0.076 & 0.081 & 0.085 & 0.081 & 0.66 & 0.65 & 0.63 & 0.65 \\

\textbf{Kinedex} &
0.052 & 0.053 & 0.055 & 0.053 & 0.70 & 0.70 & 0.69 & 0.70 &
0.045 & 0.048 & 0.052 & 0.047 & 0.80 & 0.77 & 0.75 & 0.78 &
0.017 & 0.017 & 0.017 & 0.017 & 0.60 & 0.60 & 0.58 & 0.59 \\

\textbf{Forcemimic} &
0.082 & 0.081 & 0.082 & 0.082 & 0.66 & 0.65 & 0.65 & 0.65 &
0.071 & 0.071 & 0.071 & 0.071 & 0.69 & 0.68 & 0.67 & 0.68 &
-- & -- & -- & -- & -- & -- & -- & -- \\

\textbf{Ours} &
\textbf{0.022} & \textbf{0.025} & \textbf{0.028} & \textbf{0.025} & \textbf{0.85} & \textbf{0.84} & \textbf{0.84} & \textbf{0.85} &
\textbf{0.028} & \textbf{0.032} & \textbf{0.036} & \textbf{0.030} & \textbf{0.90} & \textbf{0.89} & \textbf{0.88} & \textbf{0.89} &
\textbf{0.010} & \textbf{0.010} & \textbf{0.010} & \textbf{0.010} & \textbf{0.70} & \textbf{0.67} & \textbf{0.67} & \textbf{0.68} \\
\hline
\end{tabular}}%
\end{table*}

%% file: tables/ablation_wm.tex
\begin{table}[t]
\centering
\caption{Tactile prediction under different position, motion-history, and model settings. Lower $L_2$ and higher cosine similarity indicate better performance.}
\label{tab:ablation_wm}
\small
\setlength{\tabcolsep}{4pt}
\renewcommand{\arraystretch}{1.2}

\begin{tabular}{lccccc}
\toprule
\textbf{Settings} & \textbf{History} & \textbf{Dyn. Weight} & \textbf{Joint Gen.} & \textbf{L2} $\downarrow$ & \textbf{Cos} $\uparrow$ \\
\midrule
\multirow{4}{*}{\makecell[c]{Unseen\\Position}}
    & 3D absolute & \checkmark & \checkmark & 0.075 & 0.72 \\
    & 3D relative & \checkmark & \checkmark & 0.056 & 0.88 \\
    & 2D          & \checkmark & \checkmark & 0.042 & \textbf{0.91} \\
    & 2D + depth & \checkmark & \checkmark & \textbf{0.037} & \textbf{0.91} \\
\midrule
\multirow{4}{*}{\makecell[c]{Seen\\Position}}
    & 2D          & $\times$   & $\times$   & 0.041 & 0.90 \\
    & 2D          & $\times$   & \checkmark & 0.038 & 0.92 \\
    & 2D          & \checkmark & \checkmark & 0.035 & 0.93 \\
    & 2D + depth & \checkmark & \checkmark & \textbf{0.033} & \textbf{0.94} \\
\bottomrule
\end{tabular}
\end{table}

%% file: tables/ablation_policy.tex
\begin{table}[t]
\centering
\caption{{Adaptive Visuo-Tactile Fusion Policy ablations on wiping and peeling, all with RLTC disabled.}}
\label{tab:ablation_components_scaled}
\footnotesize
\setlength{\tabcolsep}{3pt}
\renewcommand{\arraystretch}{1.15}

\begin{tabular}{cccccccc}
\toprule
\multicolumn{5}{c}{\textbf{Component}} & \multicolumn{3}{c}{\textbf{Task Score}} \\
\cmidrule(lr){1-5} \cmidrule(lr){6-8}
\begin{tabular}[c]{@{}c@{}}Pred.\\Len.\end{tabular}
& \begin{tabular}[c]{@{}c@{}}Used as\\Contact State\end{tabular}
& LTD
& Gating
& \begin{tabular}[c]{@{}c@{}}Visual\\Gen.\end{tabular}
& Wipe
& Peel
& Avg. \\
\midrule
0 & $\times$   & $\times$   & $\times$                        & $\times$   & 0.12          & 0.06          & 0.09          \\
2 & $\times$   & $\times$   & \checkmark  & $\times$   & 0.40          & 0.26          & 0.33          \\
4 & $\times$   & $\times$   & \checkmark   & $\times$   & 0.45          & 0.30          & 0.38          \\
6 & $\times$   & $\times$   & \checkmark & $\times$   & 0.50          & 0.30          & 0.40          \\
\midrule
6 & \checkmark & $\times$   & $\times$                        & $\times$   & 0.46          & 0.28          & 0.37          \\
6 & $\times$   & \checkmark & $\times$                        & $\times$   & 0.57          & 0.36          & 0.47          \\
6 & $\times$   & \checkmark & \checkmark                      & $\times$   & 0.66          & \textbf{0.40} & 0.53          \\
6 & $\times$   & \checkmark & \checkmark                      & \checkmark & \textbf{0.70} & 0.38          & \textbf{0.54} \\
\bottomrule
\end{tabular}
\end{table}

%% file: tables/policy_detail.tex
\begin{table}[t]
\centering
\caption{Inference time of each component on RTX 4090D.}
\label{tab:policy_time}
\small
\setlength{\tabcolsep}{8pt} 
\renewcommand{\arraystretch}{1.3}
\resizebox{0.9\linewidth}{!}{
\begin{tabular}{c c c} 
\toprule
 Slow Policy & Slow Policy w/ Visual Gen. & Fast Policy \\
\midrule
 230 ms & 480 ms & 3.5 ms \\
\bottomrule
\end{tabular}
}
\end{table}

%% file: sections/6_conclusion.tex
\vspace{-3mm}

\section{Conclusion}
We address two challenges in visuo-tactile manipulation: the scarcity of large-scale, task-diverse training data and the need to model contact dynamics for closed-loop control. OmniViTac comprises $21{,}000+$ temporally aligned trajectories spanning $86$ contact-rich tasks organized into six physics-grounded interaction patterns. Our analysis of tactile signals reveals spatial locality and contact-driven dynamics, which guided OmniVTA's design. The framework learns a compact representation of local tactile deformation and predicts how contact evolves using a short-horizon visuo-tactile world model. These predictions guide multimodal fusion and action planning, while real-time tactile feedback drives high-frequency corrections. Across six interaction categories, OmniVTA ranks among the best-performing policies in most test conditions. Under perturbations, reflexive corrections help restore contact and improve task performance over open-loop execution.